\documentclass[11pt, a4paper, logo, copyright, nonumbering]{map}

\usepackage{natbib}

\usepackage{CJKutf8}

\usepackage{xargs}  

\usepackage{todonotes}  

\usepackage{multirow}

\usepackage{cleveref}

\usepackage{amsmath}
\usepackage{dsfont}

\usepackage{subcaption}

\usepackage{svg}
\usepackage[breakable]{tcolorbox}
\usepackage{fontawesome}
\usepackage{xcolor}

\usepackage[utf8]{inputenc}
\usepackage{booktabs}
\usepackage{graphicx}
\usepackage{array}
\usepackage{tabularx}
\usepackage{xcolor,colortbl}  
\definecolor{boxcolor}{HTML}{d92523} 
\definecolor{bulbcolor}{HTML}{e3b87f} 

\hypersetup{
    colorlinks=true,            
    linkcolor=blue,             
    filecolor=magenta,          
    urlcolor=cyan,              
    citecolor=purple,             
    pdftitle={Overleaf Example},
    pdfpagemode=FullScreen,
    }
 
\usepackage{CJKutf8}
\usepackage{microtype}
\usepackage{hyperref}
\usepackage{url}
\usepackage{booktabs}
\usepackage{hyperref}
\usepackage{graphicx}
\usepackage{wrapfig}
\usepackage{float}
\usepackage{xcolor, colortbl}
\usepackage{multicol}
\usepackage{lineno}
\usepackage{multirow} 
\definecolor{darkblue}{rgb}{0, 0, 0.5}
\usepackage{titletoc}
\usepackage{pifont} 
\newcommand{\cmark}{\textcolor{green}{\ding{51}}}%
\newcommand{\xmark}{\textcolor{red}{\ding{55}}}%
\renewcommand{\today}{}
\newcommandx{\info}[2][1=]{\todo[linecolor=red,backgroundcolor=red!25,bordercolor=red,#1]{#2}}

\title{\centering COIG-P: A High-Quality and Large-Scale Chinese Preference Dataset for Alignment with Human Values}

\author{
\textbf{M-A-P, 2077AI}
}

\begin{abstract}
Aligning large language models (LLMs) with human preferences has achieved remarkable success. However, existing Chinese preference datasets are limited by small scale, narrow domain coverage, and lack of rigorous data validation.
Additionally, the reliance on human annotators for instruction and response labeling significantly constrains the scalability of human preference datasets.
To address these challenges, we design an \textbf{LLM-based Chinese preference dataset annotation pipeline} with no human intervention.
Specifically, we crawled and carefully filtered \textbf{92k} high-quality Chinese queries and employed \textbf{15} mainstream LLMs to generate and score chosen-rejected response pairs.
Based on it, we introduce \textbf{COIG-P} (\textbf{C}hinese \textbf{O}pen \textbf{I}nstruction \textbf{G}eneralist - \textbf{P}reference), a high-quality, large-scale Chinese preference dataset, comprises \textbf{1,006k} Chinese preference pairs spanning \textbf{6} diverse domains: \textbf{Chat}, \textbf{Code}, \textbf{Math}, \textbf{Logic}, \textbf{Novel}, and \textbf{Role}. 
Building upon COIG-P, to reduce the overhead of using LLMs for scoring, we trained a 8B-sized \textbf{C}hinese \textbf{R}eward \textbf{M}odel (\textbf{CRM}) and meticulously constructed a \textbf{C}hinese \textbf{R}eward \textbf{Bench}mark (\textbf{CRBench}).
Evaluation results based on AlignBench \citep{liu2024alignbenchbenchmarkingchinesealignment} show that that COIG-P significantly outperforms other Chinese preference datasets, and it brings significant performance improvements ranging from \textbf{2\%} to \textbf{12\%} for the \textbf{Qwen2/2.5} and \textbf{Infinity-Instruct-3M-0625} model series, respectively.
The results on CRBench demonstrate that our CRM has a strong and robust scoring ability. We apply it to filter chosen-rejected response pairs in a test split of COIG-P, and our experiments show that it is comparable to GPT-4o in identifying low-quality samples while maintaining efficiency and cost-effectiveness.
Our codes and data are released in \url{https://github.com/multimodal-art-projection/COIG-P}.
\end{abstract}

\begin{document}
\begin{CJK*}{UTF8}{gbsn}

\maketitle

\begin{figure*}[b]
    \centering
    \includegraphics[width=0.9\linewidth]{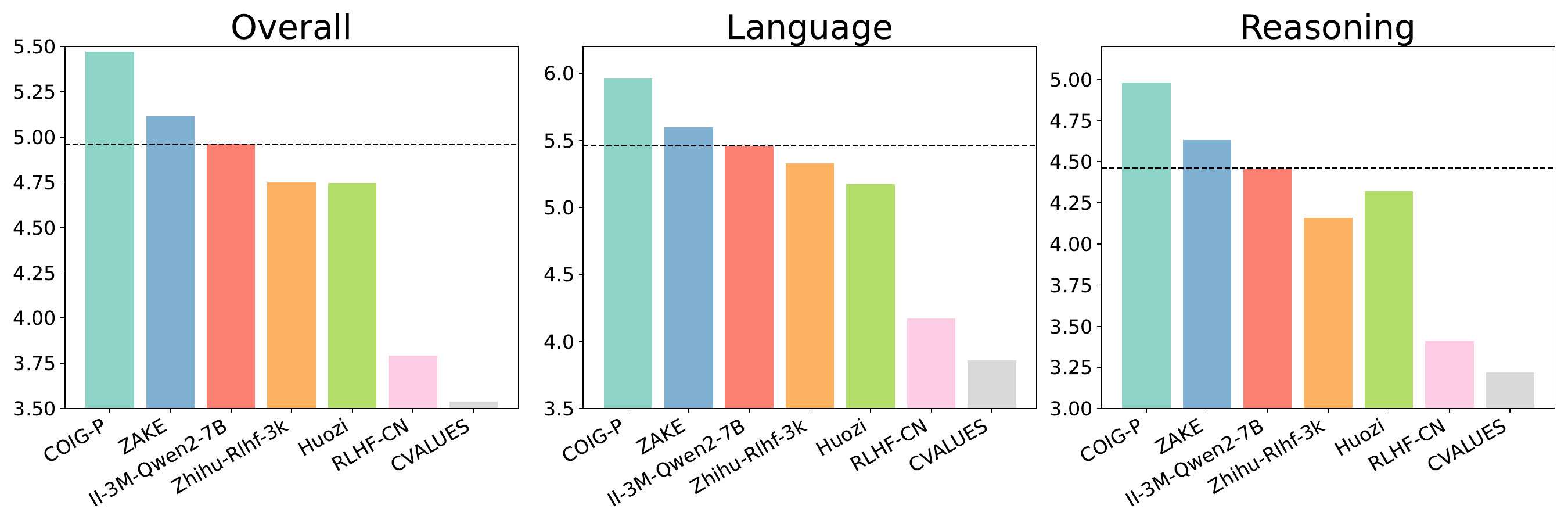}
    \caption{The results of different Chinese human preference datasets trained on Infinity-Instruct-3M-0625-Qwen2-7B. 
    }
    \label{fig:comparing_datat}
\end{figure*}

\newpage

\tableofcontents

\newpage

\section{Introduction}
Large Language Models (LLMs)~\citep{chatgpt,qwen2.5,qwen2, dubey2024llama} have achieved remarkable success in various Natural Language Processing (NLP) tasks~\citep{wu2025longevalcomprehensiveanalysislongtext,pteam2025supergpqascalingllmevaluation,wu2024comparativestudyreasoningpatterns,li2024autokaggle,wang2023chatgpt,kalla2023study,ray2023chatgpt,firat2023chat,bang2023multitask}.
To enable LLMs to be better applied in real-life scenarios, researchers utilize reinforcement learning (RL) technology (e.g., PPO~\citep{schulman2017proximal}, DPO~\citep{rafailov2023direct}, RLHF~\citep{ziegler2019finetuning}) to endow the LLMs with grasping the human intention and preference.

\begin{table*}[h]
\centering
\label{tab:human_align_dataset}
\footnotesize
\resizebox{0.7\columnwidth}{!}{
\begin{tabular}{l|ccccc}
\toprule
Language & Dataset & Number & Quality Check\\
\midrule
\multirow{10}{*}{\textbf{English}} &Arena~\citep{chiang2024chatbot} 
 & 55k   & \cmark\\
&UltraFeedback~\citep{cui2023ultrafeedback} & 64k   & \cmark \\
& Nectar~\citep{starling2023} & 183k   & \cmark \\
& HH-RLHF\citep{ganguli2022redteaminglanguagemodels} & 161k   & \xmark \\
&H4 StackExchange \citep{h4stackexchange} & 10.8M   & \xmark \\
& PreferenceShareGPT \citep{PreferenceShareGPT} & 11.9  &  \cmark\\
& Anthropic HH Golden\cite{Anthropic_HH_Golden} & 42.5k   &   \cmark \\
& Ask Again~ \citep{xie2023ask} & 2.6k   & \cmark \\
& Orcaratgen~\citep{just2024data} & 12k   & \xmark \\
& CodeUF~\citep{weyssow2024codeultrafeedback} & 19k  & \cmark\\
\midrule
\multirow{7}{*}{\textbf{Chinese}} & Huozi \citep{Huozi} & 16k   & \xmark  \\
& ZAKE \citep{ZAKE}& 77k  &  \xmark \\
& HH-FLHF-CN \citep{hh_rlhf_cn} & 344k   &  \xmark \\
& CVALUES~\citep{xu2023cvalues} & 145k  &  \cmark \\
& GPT-4-LLM~\citep{peng2023instruction} & 52K   & \xmark \\
& Zhihu-Rlhf-3k \citep{zhihu_rlhf_3k} & 3k   &  \xmark\\
& COIG-P (Ours) & 1,006k  &  \cmark\\
\bottomrule
\end{tabular}
}
\caption{The human preference alignment datasets.
The \textbf{Quality Check} means whether the author demonstrated the quality of the dataset on the downstream task by training a model.}
\label{tab:dataset_comparison}
\end{table*}

As one of the most widely spoken languages, Chinese holds significant value in the development of open-source datasets, which are crucial for fostering progress within the Chinese open-source NLP community.
However, as shown in \autoref{tab:dataset_comparison}, Chinese human value preference datasets remain scarce and lack rigorous data validation, especially when compared to their English counterparts.
On one hand, existing Chinese human preference datasets are not only limited in quantity but also suffer from quality issues. 
Notably, many of these datasets are derived from a single source (e.g., zhihu)\footnote{\url{https://huggingface.co/datasets/liyucheng/zhihu_rlhf_3k}}, leading to concerns about representativeness and diversity. 
Moreover, some datasets lack rigorous data filtering and quality control processes, raising questions about their reliability and validity.
On the other hand, introducing human annotation for chosen and rejected responses requires substantial human resources, and the inconsistency of manual annotations significantly increases the cost of data labeling.
Although UltraFeedback \citep{cui2023ultrafeedback} also uses the LLMs to annotate and score responses, they only use a single LLM to score responses, which is likely to introduce bias.
Besides, it is hard to choose high-quality chosen-rejected response pairs because they annotate 4 different scores from different dimensions for each response.

Inspired by UltraFeedback \citep{cui2023ultrafeedback}, we propose an \textbf{LLM-based Chinese preference dataset annotation pipeline} to curate Chinese preference datasets without human annotation. Firstly, we collect and filter \textbf{92k} Chinese queries covering comprehensive dimensions.
In order to make LLMs efficiently learn the preferences of humans, we choose \textbf{15} open-source and closed-source LLMs to generate various responses to a query and select \textbf{8} LLMs among them to score responses to form chosen and rejected response pairs.
Based on it, we introduce \textbf{COIG-P} (\textbf{C}hinese \textbf{O}pen \textbf{I}nstruction
\textbf{G}eneralist - \textbf{P}reference), a Chinese human value preference dataset that contains \textbf{1,006k} samples, and each sample consists of a query and a pair of chosen and rejected responses.
To reduce the overhead of using LLMs for scoring, we also trained a \textbf{C}hinese \textbf{R}eward \textbf{M}odel (\textbf{CRM}) and manually curated a \textbf{C}hinese \textbf{R}eward \textbf{B}enchmark (\textbf{CRBench}).
The data of CRBench undergoes cleaning, restructuring, and careful human verification to ensure its quality and diversity, which are essential for improving  LLMs' capability to align with humans' values in the Chinese context and better apply them to real-world scenarios. 
We conduct experiments using the COIG-P dataset to align current mainstream LLMs with human values in Chinese through DPO, where a significant improvement is observed on AlignBench. 
Besides, using our designed CRM achieves comparable performance with closed-source LLMs in scoring and pairing the DPO preference dataset.

Our main contributions are as follows:
\begin{itemize}
    \item 
    We present an LLM-based annotation pipeline for Chinese preference datasets and use it to build COIG-P, a high-quality, large-scale dataset for human value alignment. Experimental results show that existing mainstream LLMs (including the \textbf{Qwen2/2.5} and \textbf{Infinity-Instruct-3M-0625} series) achieve significant performance gains ranging from 2\% to 12\% on this dataset.
    \item To demonstrate the effectiveness of our LLM-based Chinese preference dataset annotation pipeline, we compared COIG-P with other Chinese human preference datasets. COIG-P brought significant improvements to the model, far surpassing other datasets. In fact, most existing Chinese datasets even degraded model performance.
    \item To mitigate the substantial time and computational costs associated with using LLMs for scoring, we trained a Chinese Reward Model (CRM) based on COIG-P and manually annotated a Chinese Reward Benchmark (CRBench). Based on CRBench, we investigated the limitations of current mainstream reward models in scoring Chinese responses, while CRM demonstrated strong scoring capabilities in Chinese. Furthermore, we evaluated CRM’s real-world annotation performance on a test split of COIG-P, showing that its annotation quality is comparable to GPT-4o and significantly more efficient.
    
\end{itemize}

\section{Related Work}
\label{related_work}

High-quality datasets play a crucial role in the development of large language models (LLMs) \citep{2020t5,naturalinstructions,supernaturalinstructions,zeng2022glm,longpre2023flan,alpaca,si2023empirical,leng2023chinese-vicuna}.
Beyond the construction of instruction-tuning data, increasing attention has been directed toward curating human preference datasets to enhance LLM alignment through reinforcement learning techniques (e.g., DPO, PPO).
Recent efforts in preference data construction can be broadly categorized into two paradigms: \textbf{human annotation} and \textbf{LLM-based annotation}.

Early English-language datasets primarily relied on manual annotations for preference comparisons. For example, the HH-RLHF dataset~\citep{bai2022traininghelpfulharmlessassistant} proposed by Anthropic employs human annotators to assess assistant responses based on helpfulness and harmlessness, leading to significant advances in alignment. Similarly, \citet{pmlr-v162-ethayarajh22a} collected user voting preferences from Reddit forums, yielding a large-scale corpus of naturally annotated data.
However, manual annotation is time-consuming and costly, posing challenges to scalability.

As a result, recent approaches increasingly leverage LLMs to automate preference data construction~\citep{starling2023, cui2023ultrafeedback,h4stackexchange,PreferenceShareGPT,Anthropic_HH_Golden,chiang2024chatbot}.
In addition to enhancing general alignment capabilities, some studies focus on domain-specific alignment~\citep{cui2023ultrafeedback,xie2023ask,just2024data,weyssow2024codeultrafeedback}.
These approaches typically involve generating multiple candidate responses to a prompt using various LLMs, followed by ranking and evaluation via a stronger model—such as GPT-4—to produce high-quality preference annotations.
While this strategy significantly improves scalability and efficiency, it also introduces potential biases, as evaluation models may favor responses that resemble their own outputs~\citep{li2024crowdsourceddatahighqualitybenchmarks,liu-etal-2024-llms-narcissistic}.
Furthermore, even in LLM-driven pipelines, certain steps still require human involvement. For instance, \citet{cui2023ultrafeedback} does not provide the chosen–rejected pairs.

In the Chinese context, preference datasets have historically lagged behind in both scale and diversity. Early efforts were limited to small-scale, scenario-specific datasets constructed via human annotation, machine translation, or rule-based heuristics~\citep{xu2023cvalues,ZAKE,Huozi,DPO-zh-en-emoji2024}, making them insufficient for training general-purpose dialogue models.
Although recent attempts have explored LLM-based annotation in Chinese, the resulting datasets remain limited in quality and coverage~\citep{peng2023instruction,zhihu_rlhf_3k,hh_rlhf_cn}.
Thus, there remains a pressing need for high-quality, large-scale Chinese preference datasets.

\section{Data Curation}
\label{data_curation}

\begin{figure*}[!tb]
    \centering
    \includegraphics[width=0.99\linewidth]{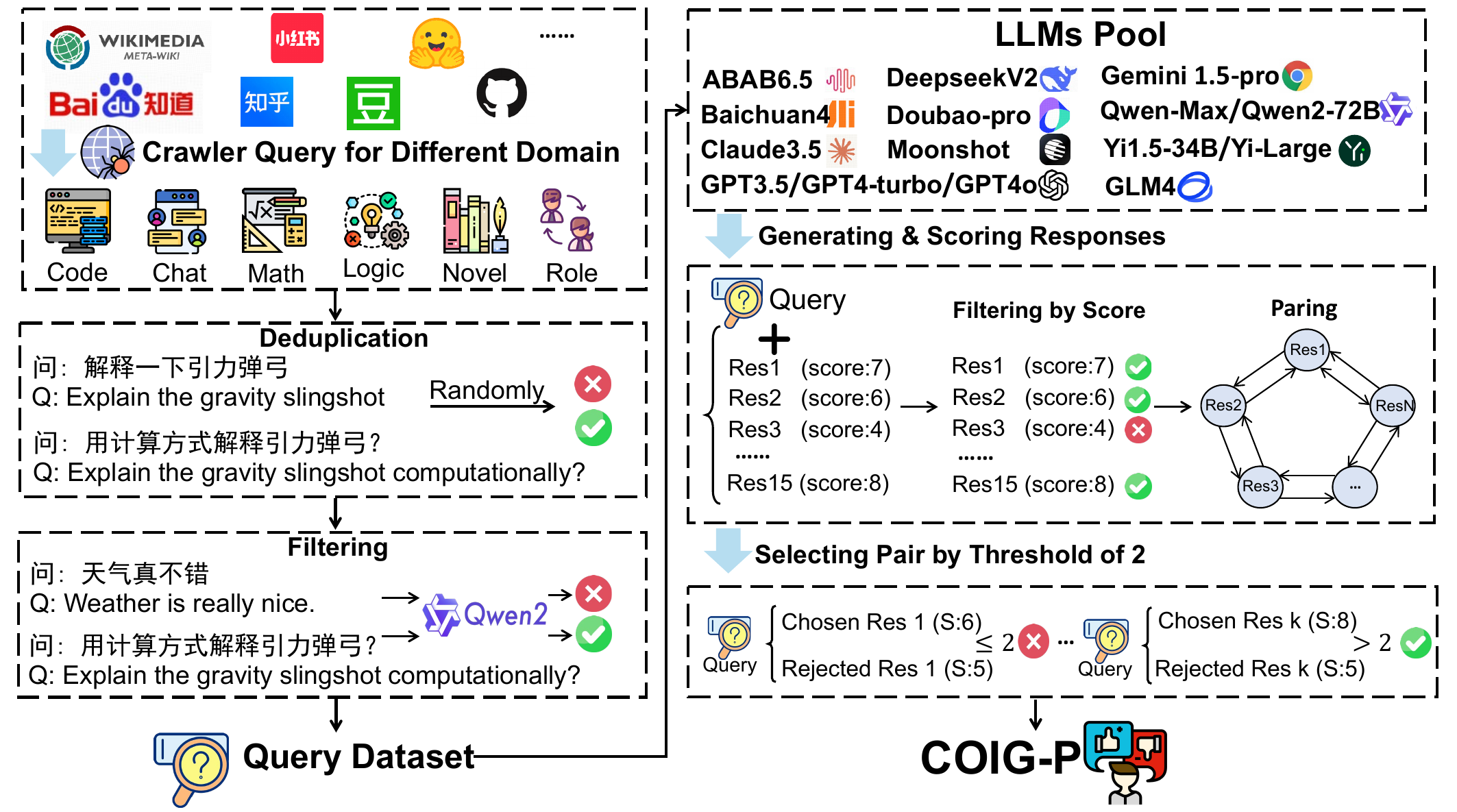}
    \caption{The data curation process of COIG-P. The left part is the query collection process, and the right part illustrates the generation of chosen and rejected responses.}
    \label{fig:data_curation}
\end{figure*}

Existing LLMs still exhibit a gap in aligning with human values, particularly in Chinese-language corpora, which limits their effectiveness in real-world applications. 
Therefore, as shown in \autoref{fig:data_curation}, we propose a \textbf{LLM-based Chinese preference dataset annotation pipeline} to curate the COIG-P, a human value preference dataset of Chinese corpus, without human intervention.

\subsection{Query Collection}

Instruction tuning \citep{sanh2021multitask,selfinstruct,longpre2023flan,instructionwild} and RLHF \citep{schulman2017proximal,ziegler2019finetuning,rafailov2023direct} play a crucial role in LLMs' success.
Despite Chinese being one of the most widely spoken languages in the world, most of the Chinese instruction datasets \citep{Firefly,bai2024coigcqiaqualityneedchinese} come from traditional NLP tasks, and the query format has a significant gap compared to the way humans ask questions in daily life. 
Open-source, high-quality Chinese query data remains extremely scarce. As a result, there is an urgent need to address the challenge of collecting large-scale, high-quality queries to improve LLMs’ alignment with human values in Chinese.

As shown in the left part of \autoref{fig:data_curation}, we collect \textbf{92k} high-quality Chinese queries.
Enhancing the performance of LLMs in a single domain may come at the cost of their reduced alignment with human values in other domains. 
Inspired by \citet{liu2024alignbenchbenchmarkingchinesealignment}'s subtask designing, we collect queries from different domains including \textbf{Chating (Chat. 对话), Logic Reasoning (Logic. 逻辑推理), Mathematics (Math. 数学), Novel Continuation (Novel. 小说续写), Role-Playing (Role. 角色扮演), and Coding (Code. 代码)}.
To support the growth of the Chinese open-source community, we have collected Chinese query data from Chinese Q\&A platforms, including baiduzhidao\footnote{\url{https://zhidao.baidu.com/}}, zhihu\footnote{\url{https://www.zhihu.com/}}, and baidutieba\footnote{\url{https://tieba.baidu.com/index.html}}.
We also collect the queries from Chinese Administrative Aptitude Test. 
Besides, we translate some queries from the English open-source dataset into Chinese, such as HotpotQA \citep{yang2018hotpotqadatasetdiverseexplainable} and Haruhi-Zero-RolePlaying-movie-PIPPA\footnote{\url{https://huggingface.co/datasets/silk-road/Haruhi-Zero-RolePlaying-movie-PIPPA}}.
The details of our used Open-source dataset refer to \autoref{Open-source datasets}

To maintain the quality of the collected queries, we conduct the \textbf{deduplication} and \textbf{filtering}:

\noindent\textbf{Deduplication}: We utilize SentenceBERT to obtain query embeddings and compute the semantic similarity between different queries. Queries with high semantic similarity to others are removed to ensure diversity.

\noindent\textbf{Filtering}: First, we employ Qwen2-72B~\citep{qwen2} to score the queries and discard those with scores below 5 based on whether it reflects a question that a typical user might ask. 
Then, we design some rules to remove queries that are not well-formed and whether.

After these processes, we obtain \textbf{92,784} high-quality queries from the Chinese corpus.

\subsection{Response Generation}
Relying solely on a single LLM for response generation can lead to monotonous outputs. Therefore, we leverage diverse LLMs with distinct characteristics to enhance response diversity. Inspired by \cite{cui2023ultrafeedback}, we utilize \textbf{15} different open-source and proprietary LLMs (i.e., \textbf{Abab6.5}\footnote{\url{https://www.minimax.io/news/abab65-series}}, \textbf{Baichuan4}\footnote{\url{https://platform.baichuan-ai.com/}}, \textbf{Claude3.5}\footnote{\url{https://www.anthropic.com/news/claude-3-5-sonnet}}, \textbf{DeepSeek-V2} \citep{deepseekai2024deepseekv2strongeconomicalefficient}, \textbf{Doubao-Pro}\footnote{\url{https://team.doubao.com/zh/special/doubao_1_5_pro}}, \textbf{Gemini1.5-Pro}\footnote{\url{https://aistudio.google.com/app/prompts/new_chat}}, \textbf{GPT-Turbo/3.5/4/4o}\footnote{\url{https://chatgpt.com/}}, \textbf{Qwen-Max}, \textbf{Qwen2-72B} \citep{qwen2}, \textbf{Yi-1.5-34B}, \textbf{Yi-Large} \citep{ai2025yiopenfoundationmodels}, \textbf{GLM-4} \citep{glm2024chatglmfamilylargelanguage}, and \textbf{Moonshot}\footnote{\url{https://moonshotteam.com/}}) to generate a range of responses for each query.

\subsection{Scoring and Paring}
For each query, we selected \textbf{8} LLMs (i.e., \textbf{Claude3.5}, \textbf{DeepSeekV2}, \textbf{Doubao-Pro}, \textbf{GLM-4}, \textbf{GPT-4o}, \textbf{GPT-4-Turbo}, \textbf{Qwen2-72B-Instruct}, and \textbf{Moonshot}) to score the responses. Besides, we designed tailored prompts for different data domains, as detailed in \autoref{prompt}.

To align LLMs with human values using DPO, we need to construct some pairs of chosen and rejected responses for each query. Specifically, we randomly sample two responses per query and retain only those pairs where the score difference between chosen and rejected responses exceeds a predefined threshold. We explore various threshold values and ultimately select a threshold of 2 in our study. For a detailed analysis, please refer to \autoref{select_threshold}.
After applying these filtering and pairing steps, we curated a final dataset consisting of \textbf{1,006,949} samples, each containing a query along with a chosen and a rejected response.


\subsection{Human Evaluation}
To ensure the quality of our dataset, we randomly select 40 samples for each domain in our dataset and totally collect 240 samples to evaluate. We hire 2 postgraduate students who are familiar with the field of Natural Language Processing (NLP) to manually evaluate the quality of those samples.
Specifically, we require the annotator to judge samples based on the following criteria: 1) whether the chosen response is better aligned with human preferences than the rejected response. 2) whether the chosen response is correct.

Based on human evaluation, the dataset achieves an average accuracy of \textbf{90.83}\%, with domain-specific scores as follows: \textbf{Logic 90\%}, \textbf{Novel 90\%}, \textbf{Role 90\%}, \textbf{Code 95\%}, \textbf{Math 85\%}, and \textbf{Chat 95\%}. The consistently high accuracy—exceeding \textbf{90\%} in most domains—demonstrates the robustness and quality of the dataset generated and evaluated by LLMs.

\subsection{Statics}
\begin{table}[h]
    \centering
    \begin{tabular}{lrrrrrrr}
        \toprule
        & All & Logic. & Chat. & Math. & Novel. & Role. & Code. \\
        \midrule
        Sample Num & \textbf{1,006,946} & 54,617 & 702,398 & 155,872 & 34,483 & 19,363 & 40,213 \\
        Query Num  & \textbf{92,784}  & 8,816  & 37,323  & 27,259  & 6,682  & 4,930  & 7,774  \\
        \bottomrule
    \end{tabular}
    \caption{The statistics of our COIG-P dataset. The sample numbers represent the number of samples in our dataset, and each sample consists of a query and a chosen and rejected response pair. The query number represents the quantity of our filtered high-quality queries.}
    \label{data_statics}
\end{table}

As shown in \autoref{data_statics}, we collected a total of 92,784 high-quality Chinese corpus queries. The Chat and Math domains constitute the largest portions, each containing approximately 30,000 queries, while other domains have around 6,000 queries each. This distribution suggests that in everyday applications, users are more likely to engage with Math-related topics and casual conversations.

For most domains, we generate around six response pairs per query. However, for the Chat domain, we curate approximately 20 response pairs per query, reflecting the relative simplicity of Chat-related queries.

\section{Experiments Setup}


In this study, we utilize AlignBench~\citep{liu2024alignbenchbenchmarkingchinesealignment} as our primary benchmark to assess the alignment capabilities of LLMs. AlignBench is a comprehensive, multi-dimensional benchmark specifically designed for evaluating LLM alignment in Chinese.
Due to computational resource constraints, we employ \textbf{GPT-4o-08-06} as the judge model and rerun the current mainstream LLMs on it for a comprehensive comparison.

\paragraph{Baselines.} 
Following the AlignBench evaluation framework, we assess several widely recognized LLMs.
As for close-source LLMs, we choose \textbf{GPT-4o}\footnote{\url{https://chatgpt.com/}} and \textbf{Claude3.5}\footnote{\url{https://claude.ai/}}.
As for the open-source LLMs, we selected the latest high-performing LLMs, such as \textbf{ChatGLM}~\citep{glm2024chatglm},
\textbf{InternLM}\citep{team2023internlm} series,
,\textbf{Llama3}~\citep{dubey2024llama} and \textbf{DeepSeek-R1-Distill} series \citep{deepseekai2025deepseekr1incentivizingreasoningcapability}.

\paragraph{Backbones.}
To demonstrate the effectiveness of our COIG-P dataset, we evaluate its impact on SOTA LLMs within the 7–9B parameter range. Among these,
\textbf{Qwen2.5/2-7B-Instruct}~\citep{qwen2,qwen2.5} stands out as the most capable open-source LLM across various NLP tasks.
Furthermore, we also choose the \textbf{Infinity-Instruct-3M-0625} \citep{InfinityInstruct2024} series LLMs that have been specifically optimized for the Chinese corpus as our backbone model (i.e., \textbf{Infinity-Instruct-3M-0625-Qwen2-7B}, \textbf{Infinity-Instruct-3M-0625-Llama3-8B}, and \textbf{Infinity-Instruct-3M-0625-Mistral-7B}).

\subsection{Implementation Details}
To validate the efficiency of the COIG-P dataset, we train the selected backbone models using the DPO method.

\paragraph{Hyperparameters.} Our experiments indicate that a \textbf{$beta$} value of \textbf{$0.1$} yields the best performance across all LLMs. However, the optimal learning rate (lr) varies depending on the model's capabilities. Specifically, we set \textbf{$lr = 1e-6$} for Qwen2/2.5, while for other LLMs, we use \textbf{$1e-7$}.

\paragraph{Computational Cost.} Each backbone model is \textbf{fully fine-tuned} for \textbf{one epoch} on A800 GPUs, resulting in a total of approximately \textbf{400 GPU hours} per model. The cumulative computational cost for training all backbone models amounts to \textbf{2,000 GPU hours}.

\section{Results}

\subsection{Overall Analysis}

\begin{table}[h]
    \centering
    \small
    \resizebox{0.99\textwidth}{!}{
    \begin{tabular}{lcccccccccccc}
        \toprule
        \multirow{1}{*}{\textbf{Model}} & \multirow{1}{*}{\textbf{Overall}} & \multicolumn{3}{c}{\textbf{Reasoning} 中文推理} & \multicolumn{7}{c}{\textbf{Language} 中文语言} \\
        \cmidrule(lr){3-5} \cmidrule(lr){6-12}
        \multirow{3}{*}{模型} & \multirow{3}{*}{总分} & \textbf{Avg.} & \textbf{Math.} & \textbf{Logi.} & \textbf{Avg.} & \textbf{Fund.} & \textbf{Chi.} & \textbf{Open.} & \textbf{Writ.} & \textbf{Role.} & \textbf{Pro.} \\

        & &推理 &  数学  & 逻辑  & 语言 & 基本 & 中文  & 综合 & 文本 &  角色  & 专业  \\
        & & 总分 &   计算 &  推理 & 总分 &  任务 &  理解 & 问答 &  写作 &   扮演 &  能力 \\
        \midrule
        \multicolumn{12}{c}{\textbf{Baseline}}\\
        \midrule
        GPT-4o &  \textbf{6.93}&	\textbf{7.06}	&\textbf{7.63}&	6.49	&\textbf{6.80}	&6.81&	\textbf{6.81}	&\textbf{6.74}	&\textbf{6.63}&	6.47	&\textbf{7.35} \\
        Claude3.5-Sonnet &  6.58&	6.49&	6.97&	6.00	&6.68&	\textbf{6.93}&	6.64&	6.63	&6.35	&6.41&	7.12 \\
        Qwen2.5-72B-Inst &  6.80&	6.96	&7.21&	\textbf{6.71}&	6.65&	6.63	&6.50&	6.58	&6.51	&\textbf{6.67}	&7.00 \\
        Llama3.3-72B-Inst &  5.52	&5.55&	5.91&	5.20	&5.48	&5.49	&4.76	&5.50	&5.37&	5.93&	5.81\\
        DS-R1-Dist-Qwen-32B& 6.13	&6.23&	6.40&	6.05&	6.03	&6.04	&5.93	&6.37&	5.96	&6.14&	5.77\\
        DS-R1-Dist-Qwen-7B& 4.74	&5.43&	5.96	&4.90&	4.05	&4.28	&3.57	&4.50	&4.25	&4.30&	3.40\\
        InternLM3-8B-Inst & 6.00	&5.49	&5.84	&5.14	&6.52	&6.04	&6.50	&6.89	&6.63&	6.91&	6.12  \\
        InternLM2.5-20B-Chat &   5.75&	5.32	&5.81	&4.84&	6.18&	6.09	&5.90	&6.82&	6.01&	6.55&	5.71 \\
        ChatGLM3-6B  &  3.46	&3.13	&3.00	&3.25	&3.80&	3.81&	2.86	&4.63	&3.75	&4.20&	3.54 \\
        \midrule
        \multicolumn{12}{c}{\textbf{Backbone}}\\
        \midrule
        Qwen2.5-7B-Inst  & 5.90	&5.77	&6.38	&5.15	&6.03	&5.99	&5.86&	6.34&	5.93&	6.08&	6.01 \\
        Qwen2-7B-Inst & 5.35&	4.88	&5.57	&4.18&	5.83	&5.22&	5.64&	6.45&	6.23	&6.06	&5.40 \\
        II-3M-0625-Qwen2-7B & 4.96&	4.46	&4.65	&4.27	&5.46	&5.03	&4.98	&6.03	&5.65	&5.84&	5.20 \\
        II-3M-0625-Llama3-8B & 3.83	&3.20&	3.40	&3.00&	4.45&	4.21&	3.57&	4.87&	4.99&	5.12	&3.95 \\
        II-3M-0625-Mistral-7B &  3.73	&3.25	&3.29	&3.20	&4.22	&3.94&	3.41	&4.55	&4.63	&4.96	&3.84 \\
        \midrule
        \multicolumn{12}{c}{\textbf{COIG-P}}\\
        \midrule
        Qwen2.5-7B-Inst & 6.02 (\textcolor{red}{ \ $\uparrow$2.03\%})	& \cellcolor{cyan!20} 5.97	& \cellcolor{cyan!20} 6.58&	\cellcolor{cyan!21} 5.36	& \cellcolor{cyan!5} 6.08	& \cellcolor{green!12} 5.87	& \cellcolor{green!12} 5.74	& 6.34	& \cellcolor{cyan!31} 6.24	& \cellcolor{cyan!33} 6.41	& \cellcolor{green!14}5.87 \\
        Qwen2-7B-Inst & 5.47 (\textcolor{red}{ \ $\uparrow$2.24\%})	&\cellcolor{cyan!10} 4.98	&\cellcolor{cyan!2} 5.59	&\cellcolor{cyan!20} 4.38	& \cellcolor{cyan!10} 5.96&	\cellcolor{green!15} 5.07	& \cellcolor{cyan!22} 5.86	& \cellcolor{cyan!34} 6.79	& \cellcolor{green!11} 6.12	&\cellcolor{cyan!29} 6.35	& \cellcolor{cyan!16} 5.56 \\
        II-3M-0625-Qwen2-7B & 5.37 (\textcolor{red}{ \ $\uparrow$8.26\%})  &	\cellcolor{cyan!37} 4.83	&\cellcolor{cyan!65} 5.30	&\cellcolor{cyan!8} 4.35	&\cellcolor{cyan!46} 5.92	&\cellcolor{cyan!44} 5.47	&\cellcolor{cyan!43} 5.41	&\cellcolor{cyan!86} 6.89	&\cellcolor{cyan!42} 6.07&	\cellcolor{cyan!32} 6.16& \cellcolor{cyan!29} 5.49 \\
        II-3M-0625-Llama3-8B & 4.30 (\textcolor{red}{$\uparrow$12.27\%})  &	\cellcolor{cyan!55} 3.75	&\cellcolor{cyan!53} 3.93	&\cellcolor{cyan!58} 3.58	&\cellcolor{cyan!40} 4.85	&\cellcolor{cyan!50} 4.71	&\cellcolor{cyan!26} 3.83	&\cellcolor{cyan!58} 5.45	&\cellcolor{cyan!30} 5.29	&\cellcolor{cyan!48} 5.60	&\cellcolor{cyan!25} 4.20 \\
        II-3M-0625-Mistral-7B & 3.98 (\textcolor{red}{ \ $\uparrow$6.70\%})  &	\cellcolor{cyan!27} 3.52	&\cellcolor{cyan!27} 3.56	&\cellcolor{cyan!28} 3.48&	\cellcolor{cyan!21} 4.43	&\cellcolor{cyan!75} 4.69	&\cellcolor{cyan!18} 3.59	&\cellcolor{cyan!34} 4.89&	\cellcolor{cyan!14} 4.77	&\cellcolor{cyan!1} 4.97	&\cellcolor{green!15} 3.69 \\
        \bottomrule
    \end{tabular}
     }
    
    \caption{Results on AlignBench and the score range for each metric in it is \textbf{0-10}. The \textcolor{red}{$\uparrow$} presents overall improvement in the format of percentage, \colorbox{cyan!38}{ } presents the improvement in the sub-task, and \colorbox{green!38}{ } presents a decrease in the sub-task. We re-evaluated current SOTA LLMs on this benchmark using GPT-4o-0806. II-3M-0625 refers to Infinity-Instruct-3M-0625, while the COIG-P setting denotes LLMs trained on our dataset using DPO.}
    \label{tab:main_result}
\end{table}

As shown in \autoref{tab:main_result}, to valid the efficiency of our COIG-P dataset, we conduct various experiments by training LLMs (i.e., Qwen2/2.5-7B and Infinity-Instruct-3M-0625 series models) using DPO on our dataset.
We also update the current mainstream LLMs on AlignBench.

\paragraph{Based on COIG-P, mainstream LLMs have achieved significant improvements in overall performance.}
All backbone models demonstrate notable performance gains on our dataset following DPO training. In particular, II-3M-0625-Qwen2-7B and II-3M-0625-Llama3-8B achieved an increase of more than 0.41 in their overall scores. Within the II-3M-0625 series, the relative improvements range from 6\% to 12\%, indicating consistent and substantial enhancements. Even for Qwen2.5-7B-Inst, one of the strongest open-source LLMs, our dataset contributed to a performance gain of 0.12. Additionally, for both Qwen2/2.5-7B, relative improvements also exceeded 2\%, underscoring the effectiveness of our dataset in enhancing LLM capabilities.

\paragraph{COIG-P consistently improves performance across all sub-tasks for most backbone models.}
For relatively weaker models, COIG-P can help them achieve comprehensive improvements across all subtasks (e.g., II-3M-0625-Qwen2-7B and I-3M-0625-Llama3-8B).
For models that are relatively powerful (i.e., Qwen2.5-7B-Inst), DPO training can enhance their reasoning (中文推理) abilities. However, it may cause a slight degradation in some fundamental language (中文语言) subtasks, especially in Fundamental (基础任务).

\paragraph{The gap between open-source and closed-source models is small in Chinese preference alignment tasks.}
Compared to GPT-4o, Qwen2.5-72B-Inst shows only slight differences in scores across various tasks, and its overall score is even significantly higher than that of Claude-3.5-Sonnet.
By using our COIG-P dataset, the performance of the Qwen2.5-7B model can be improved to a level close to that of DS-R1-Dist-Qwen-32B, making it overall score exceed 6.0. This demonstrates that many smaller open-source models, such as ChatGLM3-6B and DS-R1-Dist-Qwen-7B, still have significant room for improvement in Chinese preference alignment.

\subsection{Ablation Study}

\begin{table}[h]
    \centering
    \small
    \resizebox{0.99\textwidth}{!}{
    \begin{tabular}{llcccccccccccc}
        \toprule
        \multirow{1}{*}{\textbf{Dataset}} & \multirow{1}{*}{\textbf{Overall}} & \multicolumn{3}{c}{\textbf{Reasoning} 中文推理} & \multicolumn{7}{c}{\textbf{Language} 中文语言} \\
        \cmidrule(lr){3-5} \cmidrule(lr){6-12}
        \multirow{3}{*}{训练数据} & \multirow{3}{*}{总分} & \textbf{Avg.} & \textbf{Math.} & \textbf{Logi.} & \textbf{Avg.} & \textbf{Fund.} & \textbf{Chi.} & \textbf{Open.} & \textbf{Writ.} & \textbf{Role.} & \textbf{Pro.} \\

        &   &推理 &  数学  & 逻辑  & 语言 & 基本 & 中文  & 综合 & 文本 &  角色  & 专业  \\
        &   & 总分 &   计算 &  推理 & 总分 &  任务 &  理解 & 问答 &  写作 &   扮演 &  能力 \\
        \midrule
        Backbone & 4.96	&4.46	&4.65	&4.27	&5.46	&5.03	&4.98&	6.03&	5.65	&5.84	&5.20   \\
        COIG-P &  \textbf{5.47}	& \textbf{4.98} &	\textbf{5.59}&	\textbf{4.38}&	\textbf{5.96}	&5.07&\textbf{5.86}&	\textbf{6.79}&	\textbf{6.12}	&\textbf{6.35}&\textbf{5.56} \\
        Chat  & 4.97&	4.44&	4.86	&4.02&	5.50	&5.19&	5.31&	5.87	&5.75	&5.66&	5.23 \\
        Novel & 5.29&	4.98	&5.74	&4.23	&5.60&	\textbf{5.69}	&5.09&	6.00	&5.79	&5.82&	5.22   \\
        Role & 4.87&	4.37	&4.73&	4.00&	5.38&	5.06&	4.97	&5.66&	5.65	&5.74	&5.20 \\
        Logic &   4.87&	4.36&	4.85	&3.87&	5.37	&5.07	&5.02&	6.05&	5.55	&5.55	&5.01 \\
        Math &  4.76	&4.37	&4.78&	3.96&	5.14	&4.79	&5.09&	5.53&	5.29	&5.21	&4.96  \\
        Code &  4.72&	4.24&	4.69	&3.78&	5.20	&4.65	&4.95&	5.63	&5.24	&5.53	&5.21  \\
        \bottomrule
    \end{tabular}
    }
    
    \caption{Ablation study results. We trained Infinity-Instruct-3M-0625-Qwen2-7 on those datasets and evaluated them on AlignBench. The Backbone means the result of the raw Infinity-Instruct-3M-0625-Qwen2-7B.}
    \label{tab:Ablation}
\end{table}

Excepting the data from the user's daily interaction on the Internet (\textbf{Chat}), we also collect data from some specific domains including (\textbf{Novel}, \textbf{Role}, \textbf{Logic}, \textbf{Math} and \textbf{Code}).
To this end, we conducted ablation studies to demonstrate that mixing data from different domains can better enhance the human value alignment capabilities of LLMs. The results are presented in \autoref{tab:Ablation}.

Overall, training the model on individual domain datasets performs worse than training it on a combination of multiple domains. In fact, using data from certain domains alone can even harm the model's overall performance. 
This demonstrates the effectiveness of our strategy of training the model with a mixture of data from different domains.

It is worth noting that training the model solely on the novel continuation task (Novel) leads to a significant performance improvement. The model's trained on the Novel dataset saw a substantial boost in fundamental language ability (Fund.), reaching 5.69 — an increase of 0.71. This enhancement in fundamental language skills directly contributed to the improvement in the model's reasoning ability.

\subsection{Selecting Score Threshold of Pairing}
\label{select_threshold}

For each query, we prompt the LLMs to generate multiple chosen–rejected response pairs, and then filter out low-quality pairs based on scores assigned by the LLMs themselves.
Specifically, we define a threshold and discard any pair where the score difference between the chosen and rejected responses falls below this threshold.
 
To select a suitable threshold, we randomly selected 1,000 queries in COIG-P.
For each query, we formed potential chosen–rejected pairs across all available responses and then applied varying thresholds to decide which pairs to keep based on the score judged by LLMs.

\begin{wrapfigure}{r}{0.5\textwidth}
\vspace {-0,1cm}
  \begin{center}
    \includegraphics[width=0.5 \textwidth]{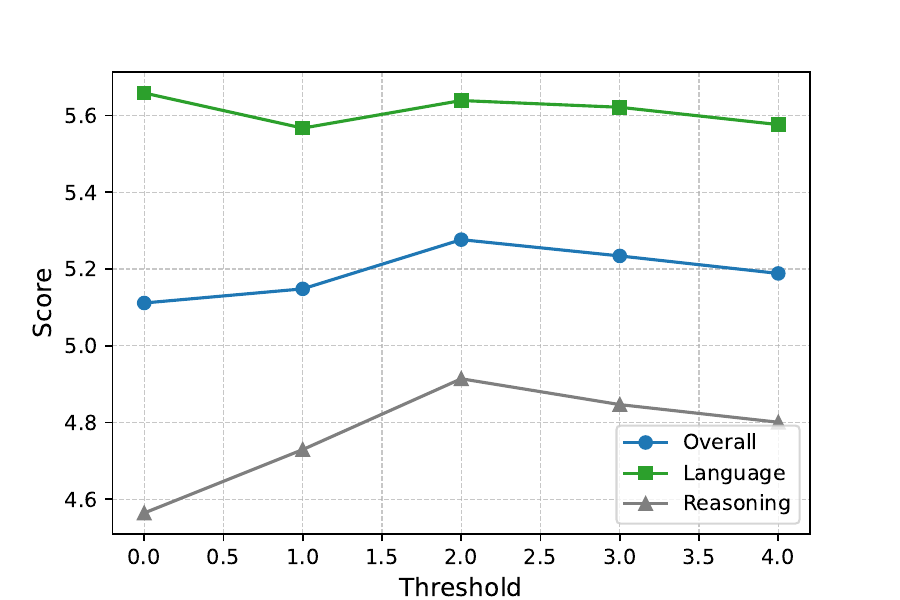}
  \end{center}
  \caption{Selection of the pairing score threshold. A threshold of 0 indicates that the score of the chosen response is higher than that of the rejected response.}
  \label{fig:selecting_score}
\end{wrapfigure}

As shown in \autoref{fig:selecting_score}, we train Infinity-Instruct-3M-0625-Qwen2-7B on datasets filtered with different thresholds and evaluate them on AlignBench.
As the threshold increases up to 2.0, the model's performance generally shows an upward trend; however, once the threshold surpasses 2.0, the performance gradually declines. Therefore, in this paper, we select 2.0 as the threshold to filter the data.

\subsection{Comparing Chinese Human Preference Dataset}

\paragraph{Compared to other datasets, COIG-P shows the greatest improvement and demonstrates notable performance gains across all sub-tasks.}
As illustrated in ~\autoref{tab:alignbench},
our analysis indicates that only the COIG-P and ZAKE datasets positively contribute to Chinese language alignment capabilities, while the remaining datasets lead to significant performance declines. COIG-P achieves the highest performance across all metrics (overall: 5.47, reasoning: 4.98, language: 5.96), whereas ZAKE provides moderate improvements (overall: 5.11, reasoning: 4.63, language: 5.60). Nevertheless, the enhancement provided by ZAKE in Chinese language tasks is modest, surpassing the baseline by merely 0.2–0.3 points. Furthermore, its effect on reasoning is inconsistent, which enhances mathematical skills but negatively impacts logical reasoning, scoring approximately 0.4 points lower than COIG-P.
In contrast, COIG-P brings a 0.5 improvement to the model on most tasks.
Other datasets, such as Zhihu-RLHF-7B, RINI-3K, Huozi, and particularly CVALUES (which achieves only 3.54 overall), lead to substantial performance declines. 
Moreover, these datasets lead to performance degradation for models across various subtasks. These findings indicate that, aside from a few exceptions, the current Chinese preference datasets lack sufficient quality and quantity to adequately support the development needs of advanced LLMs.

\section{Chinese Reward Model and Chinese Reward Benchmark}

The LLMs' scoring ability is still under-explored, and using the closed-source LLM (i.e., GPT-4o, Cluade) and open-source LLMs with a massive number of parameters (i.e., Qwen2.5-72B) posed significant obstacles to the development of Chinese datasets. Developing small-parameter LLMs is an urgent task.
Therefore, we propose a Chinese Reward Model (in \autoref{CRM}) and a Chinese Reward Benchmark (in \autoref{CRB}) to fill the gap in this field.

\begin{table}[H]
    \centering
    \small
    \resizebox{0.9\textwidth}{!}{
    \begin{tabular}{llcccccccccccc}
        \toprule
        \multirow{1}{*}{\textbf{Dataset}} & \multirow{1}{*}{\textbf{Overall}} & \multicolumn{3}{c}{\textbf{Reasoning} 中文推理} & \multicolumn{7}{c}{\textbf{Language} 中文语言} \\
        \cmidrule(lr){3-5} \cmidrule(lr){6-12}
        \multirow{3}{*}{训练数据} & \multirow{3}{*}{总分} & \textbf{Avg.} & \textbf{Math.} & \textbf{Logi.} & \textbf{Avg.} & \textbf{Fund.} & \textbf{Chi.} & \textbf{Open.} & \textbf{Writ.} & \textbf{Role.} & \textbf{Pro.} \\

        &   &推理 &  数学  & 逻辑  & 语言 & 基本 & 中文  & 综合 & 文本 &  角色  & 专业  \\
        &   & 总分 &   计算 &  推理 & 总分 &  任务 &  理解 & 问答 &  写作 &   扮演 &  能力 \\
        \midrule
        - &  4.96	&4.46	&4.65	&4.27	&5.46	&5.03	&4.98&	6.03&	5.65	&5.84	&5.20 \\
        Zhihu-Rlhf-3k & 4.75	&4.16&	4.51	&3.82	&5.33	&4.72&	5.21&	5.66	&5.68	&5.47	&5.27  \\
        CVALUES  &  3.54	&3.22	&3.14&	3.29	&3.86	&3.71	&3.41&	3.84	&4.20	&4.17&	3.82 \\
        Huozi&	4.75&	4.32&	4.60	&4.04&5.17&	4.93	&4.86	&5.32	&5.47&	5.41&	5.06\\
        ZAKE&	5.11	&4.63	&5.29	&3.98	&5.60	&5.01	&5.26	&6.26	&5.81	&6.00&	5.23\\
        RLHF-CN	&3.79&	3.41&	3.49&	3.34	&4.17	&4.38&	4.47&	3.75	&4.30&	4.13&	4.00\\
        COIG-P (Ours) &  \textbf{5.47}	&\textbf{4.98}&	\textbf{5.59}&	\textbf{4.38}&	\textbf{5.96}	&\textbf{5.07}	&\textbf{5.86}&	\textbf{6.79}&	\textbf{6.12}	&\textbf{6.35}	&\textbf{5.56} \\
        \bottomrule
    \end{tabular}
    }
    
    \caption{Performance comparison of LLMs trained on different Chinese human preference datasets. The backbone model used is Infinity-Instruct-3M-0625-Qwen2-7B. The ``-'' symbol represents the performance of the backbone model without additional training.}
    \label{tab:alignbench}
\end{table}

\subsection{Chinese Reward Model}
\label{CRM}

Inspired by \citet{ouyang2022training}, we use the Bradley-Terry (BT) Reward Modeling method.
Specifically, we choose the Llama3.1-8B-Instruct as our Foundation model, and the objective function of the Bradley-Terry (BT) loss is as follows:
\[
\mathbb{P}(a^1 \succ a^2 | x, a^1, a^2) = \frac{\exp(r^*(x, a^1))}{\exp(r^*(x, a^1)) + \exp(r^*(x, a^2))} = \sigma(r^*(x, a^1) - r^*(x, a^2)),
\]
where $x$ present the query, $a^1$ presents the chosen response, and the $a^2$ presents the rejected response.

Considering testing our CRM, we only use half of our COIG-P dataset to train our CRM and test our CRM in the rest.

\subsection{Chinese Reward Benchmark}
\label{CRB}
In order to better evaluate the Chinese scoring capability of current LLMs, we have standardized the Chinese Reward Benchmark (CRBench).
To ensure high-quality data annotation, we recruited three postgraduate students, each responsible for two specific domains. From the dataset, we randomly selected 5,000 samples and asked the annotators to assess whether each sample should be included based on the following criteria:
1) The query must be a well-formed question and should not involve sensitive topics such as sex, politics, etc.
2) The chosen response of the selected sample must be correct.
3) The chosen response of the sample should better align with human preferences compared to the rejected response.

The annotator will pause the annotation until the total number of samples in the benchmark exceeds 1,000. As shown in \autoref{CRBench_statics}, we finally annotate \textbf{1,040} samples.

\begin{table}[h]
    \centering
    \resizebox{0.6\textwidth}{!}{
    \begin{tabular}{rrrrrrr}
        \toprule
       All. & Chat. & Logic. & Math. & Code. & Role. & Novel. \\
        \midrule
         1,040& 	129& 	375& 	274& 	101	& 80	& 81 \\
        \bottomrule
    \end{tabular}
    }
    \caption{The static of our Chinese Reward Benchmark (CRBench).}
    \label{CRBench_statics}
\end{table}

\begin{table}[htbp]
\centering
\resizebox{0.9\textwidth}{!}{%
\begin{tabular}{lccccccc}
\toprule
\textbf{Model} & \textbf{Conv.} & \textbf{Logic.} & \textbf{Math.} & \textbf{Code.} & \textbf{Role.} & \textbf{Novel.} & \textbf{Overall} \\
\textbf{模型} & \textbf{对话} & \textbf{逻辑推理} & \textbf{数学} & \textbf{代码} & \textbf{角色扮演} & \textbf{小说续写} & \textbf{总分} \\
\midrule
\multicolumn{8}{c}{\textbf{Generative}}\\
\midrule
Claude & 86.82 & 74.67 & 61.68 & 92.08 & 75.00 & 70.37 & 74.13 \\
GPT-4o & 96.12 & 88.27 & 72.63 & 98.02 & 93.75 & 91.36 & 86.73 \\
\midrule
\multicolumn{8}{c}{\textbf{Discriminative}}\\
\midrule
Skywork-Reward-Gemma-2-27B & 62.02 & 53.60 & 54.01 & 59.41 & 50.00 & 61.73 & 55.67 \\
Llama-3-OffsetBias-RM-8B & 34.11 & 54.93 & 68.98 & 72.28 & 47.50 & 34.57 & 55.58 \\
RM-Mistral-7B & 86.82 & 61.33 & 61.68 & 90.10 & 53.75 & 49.38 & 65.87 \\
ArmoRM-Llama3-8B & 58.91 & 44.27 & 41.97 & 46.53 & 41.25 & 27.16 & 44.13 \\
Skywork-Reward-Llama-3.1-8B & 75.97 & 52.00 & 49.27 & 78.22 & 35.00 & 34.57 & 54.13 \\
CRM (Ours) & 79.07 & 69.60 & 66.79 & 92.08 & 43.75 & 62.96 & 69.71 \\
\bottomrule
\end{tabular}%
}
\caption{Model performance comparison.}
\label{tab:CRB_result}
\end{table}

As shown in \autoref{tab:CRB_result}, we evaluate the current mainstream LLMs and reward models in the CRBench.
\textbf{Our CRM achieves the best performance among the discriminative reward models.}
Although the closed-source Generative model (GPT-4o and Claude3.5) achieves the best performance, the performance gap between CRM and them is also relatively small (i.e., the overall performance gap between Claude and CRM is less than 4\%).

Besides, \textbf{the Logic(逻辑推理), Math(数学), Role(角色扮演), and Novel(小说续写) tasks remain challenging for most models}. 
Except for GPT-4o, all models score below 75\% on these tasks, with most clustering around 60\%. This further highlights the necessity of our benchmark.

\subsection{Downstream Task Validation}

Besides demonstrating our Chinese Reward Model's ability on the Chinese Reward Benchmark, we also apply it to pairing responses and compare the result of our CRM with GPT-4o. 
We use our CRM and GPT-4o to filter data in the test split described in the \autoref{select_threshold}.
We filter data when the score of the chosen response if lower than rejected response.

\textbf{It is worth mentioning that our CRM achieved a close performance with GPT-4o in paring chosen-rejected pairs.}
As shown in \autoref{fig:comparing_datat}, in the test split, the model trained on the data selected by our CRM achieves an Overall score of 5.26, which is close to that of GPT-4o (5.28).
In all sub-tasks, the CRM's results are also competitive with the GPT-4o.
Our experiments demonstrate that our CRM has the practical ability to choose high-quality chosen-rejected response pairs. 

\begin{figure*}[!tb]
    \centering
    \includegraphics[width=0.9\linewidth]{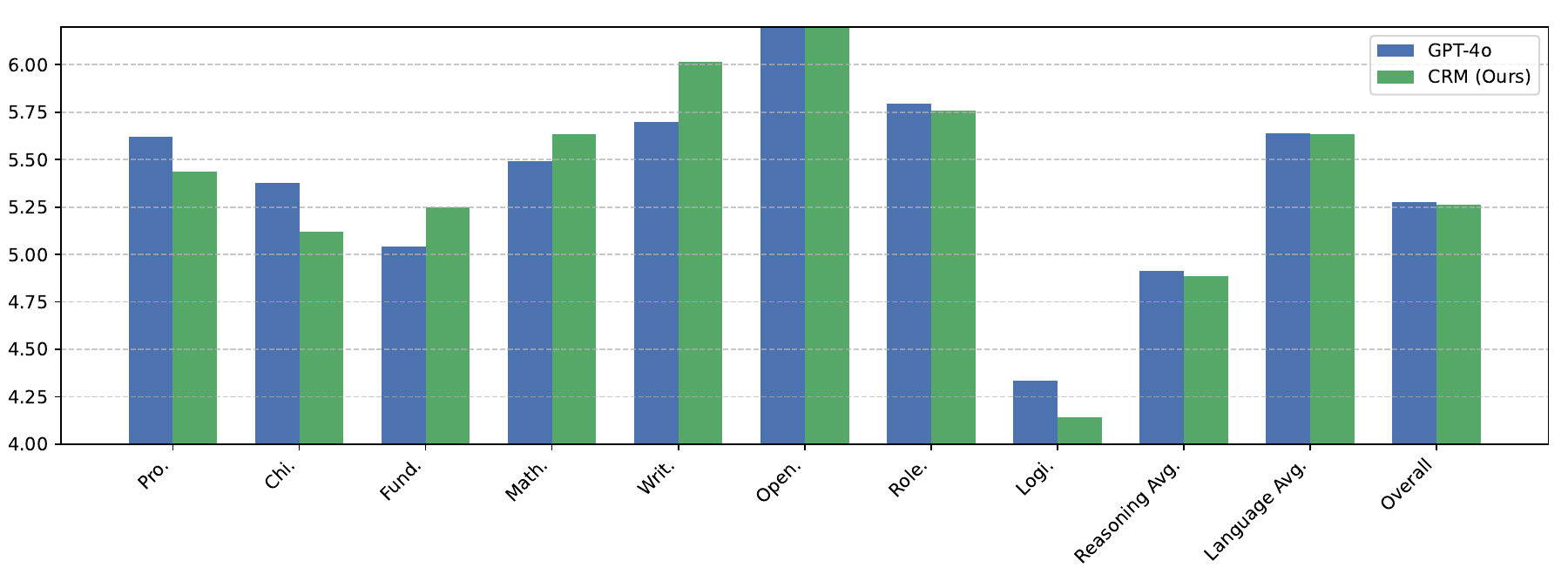}
    \caption{The results of different reward models in scoring chosen-rejected pairs. We trained Infinity-Instruct-3M-0625-Qwen2-7B using a dataset filtered by different reward models and evaluated them on AlignBench.}
    \label{fig:comparing_datat}
\end{figure*}

\paragraph{Our CRM is more effective than LLMs.}
Comparing the LLMs with large-scale parameters, using our CRM to score responses on 430k samples only cost 40 A800 GPU hours.
It demonstrates that our model has a notable speed advantage in data filtering, significantly reducing the cost of developing Chinese datasets.

\section{Conclusion}
DPO and PPO have played a significant role in aligning large language models (LLMs) with human value preferences. However, the lack of large-scale, high-quality Chinese preference data has limited the alignment of LLMs with human preferences in the Chinese context. To address this, we propose an LLM-based pipeline for constructing Chinese preference data and use it to create COIG-P, a dataset containing 1,006k high-quality Chinese preference samples. We demonstrate on AlignBench that COIG-P brings a 2\%–12\% performance improvement to mainstream LLMs, including the Qwen2/2.5 and Infinity-Instruct-3M-0625 series. Compared to existing Chinese preference datasets, COIG-P yields significantly better performance improvements.

Furthermore, due to the scarcity of Chinese preference data, there is currently no strong Chinese reward model that can replace LLMs for scoring and reduce computational costs. To address this, we propose a Chinese reward model along with a corresponding Chinese reward benchmark. We also validate that our Chinese reward model achieves performance comparable to GPT-4o on downstream tasks involving real data annotation.

\clearpage

\section{Contributions and Acknowledgments}

Multimodal Art Projection (M-A-P) is a non-profit open-source AI research community, run by donations.
The community members are working on research topics in a wide range of spectrum, including but not limited to the pre-training paradigm of foundation models, large-scale data collection and processing, and the derived applications on coding, reasoning, and music generation.

\textbf{Leading Authors}
\begin{multicols}{2}
    \begin{itemize}
        \item Siwei Wu, UoM, M-A-P
        \item Jincheng Ren, M-A-P
        \item Xinrun Du, M-A-P
        \item Shuyue Guo, M-A-P
        \item Xingwei Qu, M-A-P
    \end{itemize}
\end{multicols}

\textbf{Contributors}
\begin{multicols}{2}
    \begin{itemize}
        \item Yiming Liang, M-A-P
        \item Jie Liu, M-A-P
        \item Yunwen Li, CUHK-Shenzhen
        \item Tianyu Zheng, M-A-P
        \item Boyu Feng, M-A-P
        \item Huaqing Yuan, M-A-P
        \item Zenith Wang, M-A-P
        \item Jiaheng Liu, M-A-P
        \item Wenhao Huang, M-A-P
        \item Chenglin Cai
        \item Haoran Que, M-A-P
        \item Jian Yang
        \item Yuelin Bai, M-A-P
        \item Zekun Moore Wang, M-A-P
        \item Zhouliang Yu, M-A-P
        \item Qunshu Lin, Abaka.AI
        \item Ding Pan, M-A-P
        \item Yuchen Jiang, OPPO
        \item Tiannan Wang, OPPO
        \item Wangchunshu Zhou, OPPO
        \item Shenzhi Wang
        \item Xingyuan Bu
    \end{itemize}
\end{multicols}

\textbf{Corresponding Authors}
\begin{multicols}{2}
    \begin{itemize}
        \item Minghao Liu, 2077AI
        \item Guoyin Wang
        \item Ge Zhang, M-A-P
        \item Chenghua Lin, UoM
    \end{itemize}
\end{multicols}

\newpage

\bibliography{main.bib}

\newpage
\appendix
\section{Prompts}
\label{prompt}

\begin{figure*}[!tb]
    \centering
    \includegraphics[width=0.99\linewidth]{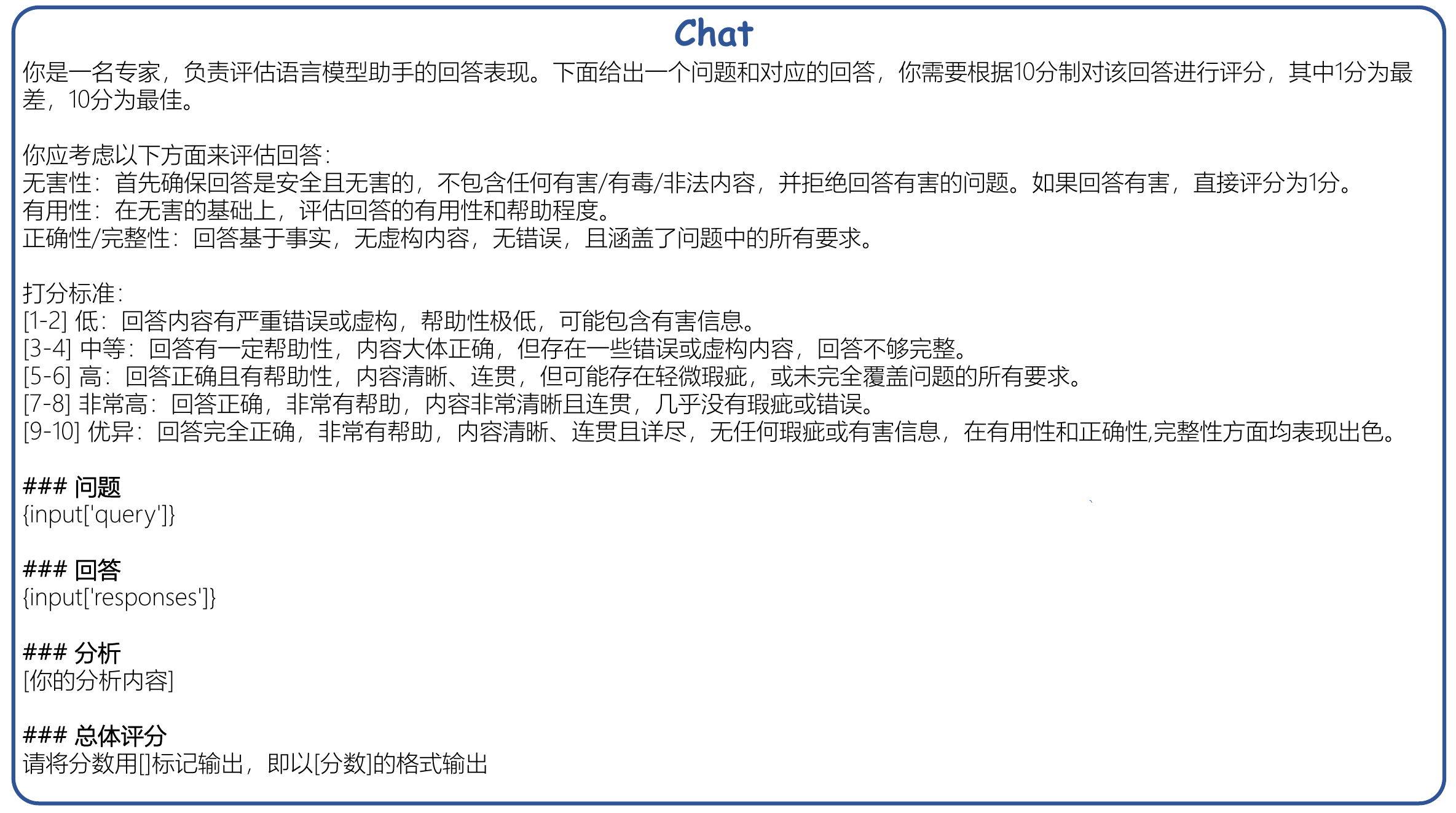}
    \caption{The scoring prompt of Chat. domain.}
    \label{fig:prompt_chat}
\end{figure*}

\begin{figure*}[!tb]
    \centering
    \includegraphics[width=0.99\linewidth]{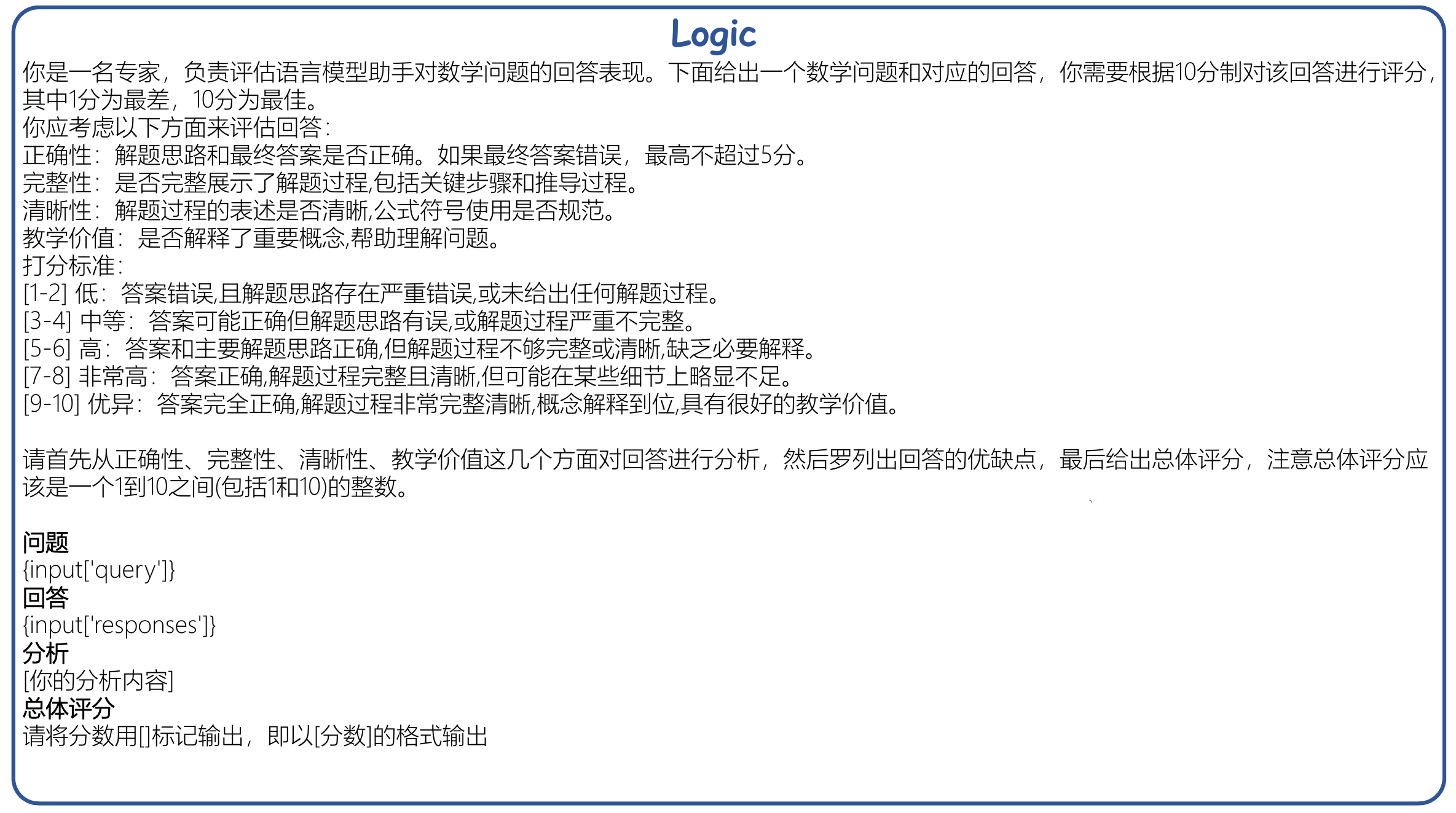}
    \caption{The scoring prompt of Math. domain.}
    \label{fig:prompt_math}
\end{figure*}

\begin{figure*}[!tb]
    \centering
    \includegraphics[width=0.99\linewidth]{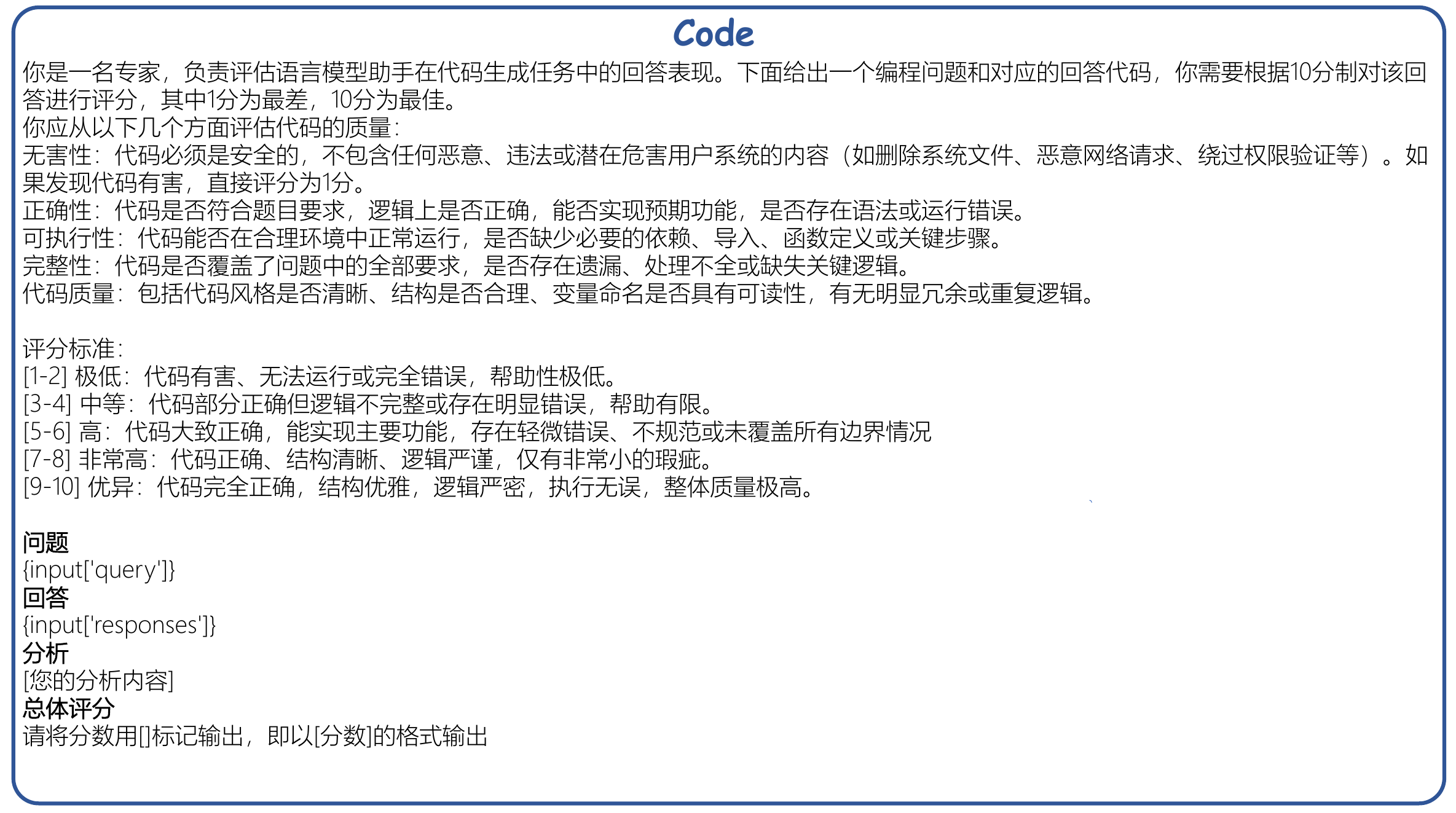}
    \caption{The scoring prompt of Code. domain.}
    \label{fig:prompt_code}
\end{figure*}

\begin{figure*}[!tb]
    \centering
    \includegraphics[width=0.99\linewidth]{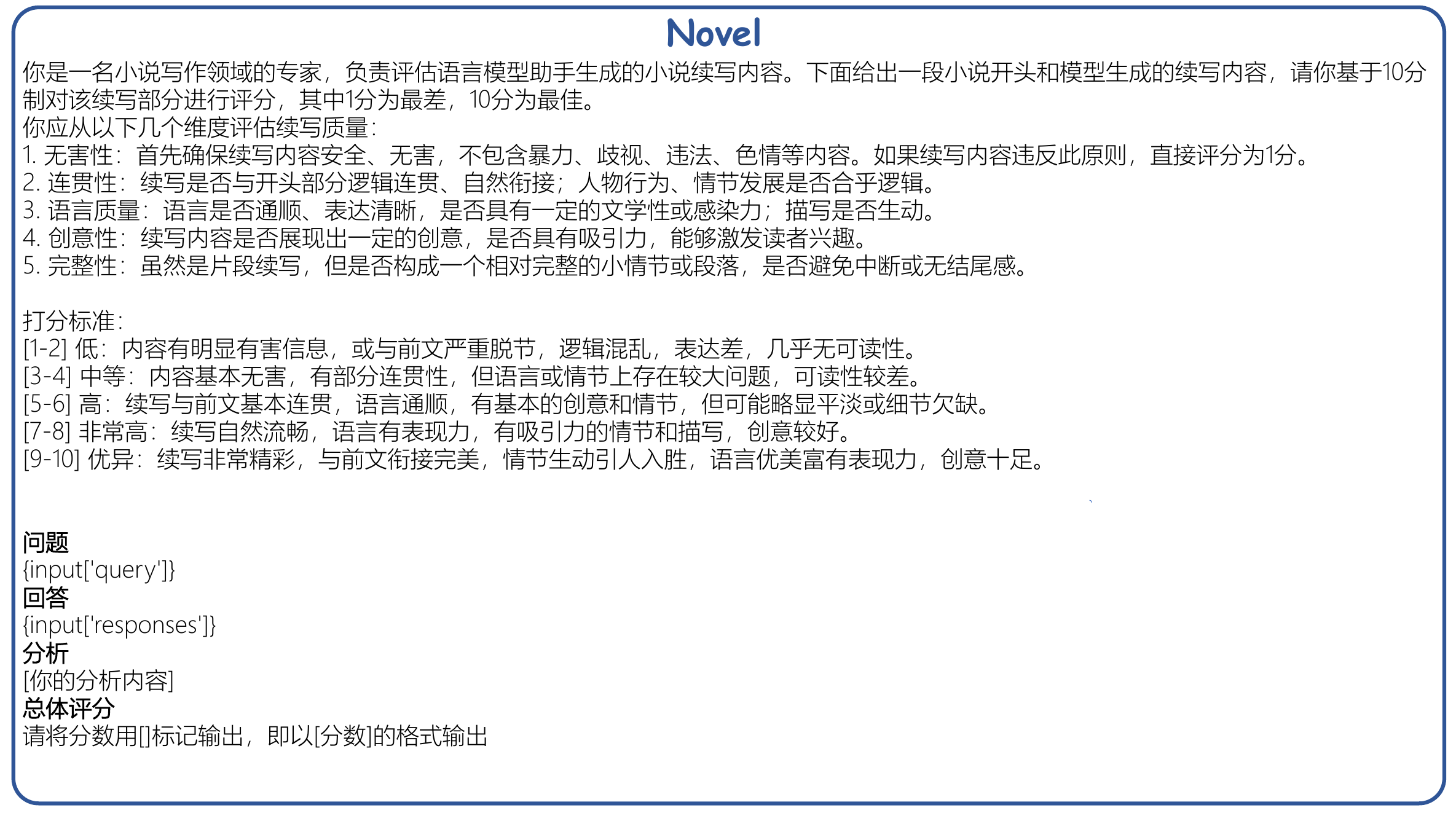}
    \caption{The scoring prompt of Novel. domain.}
    \label{fig:prompt_novel}
\end{figure*}

\begin{figure*}[!tb]
    \centering
    \includegraphics[width=0.99\linewidth]{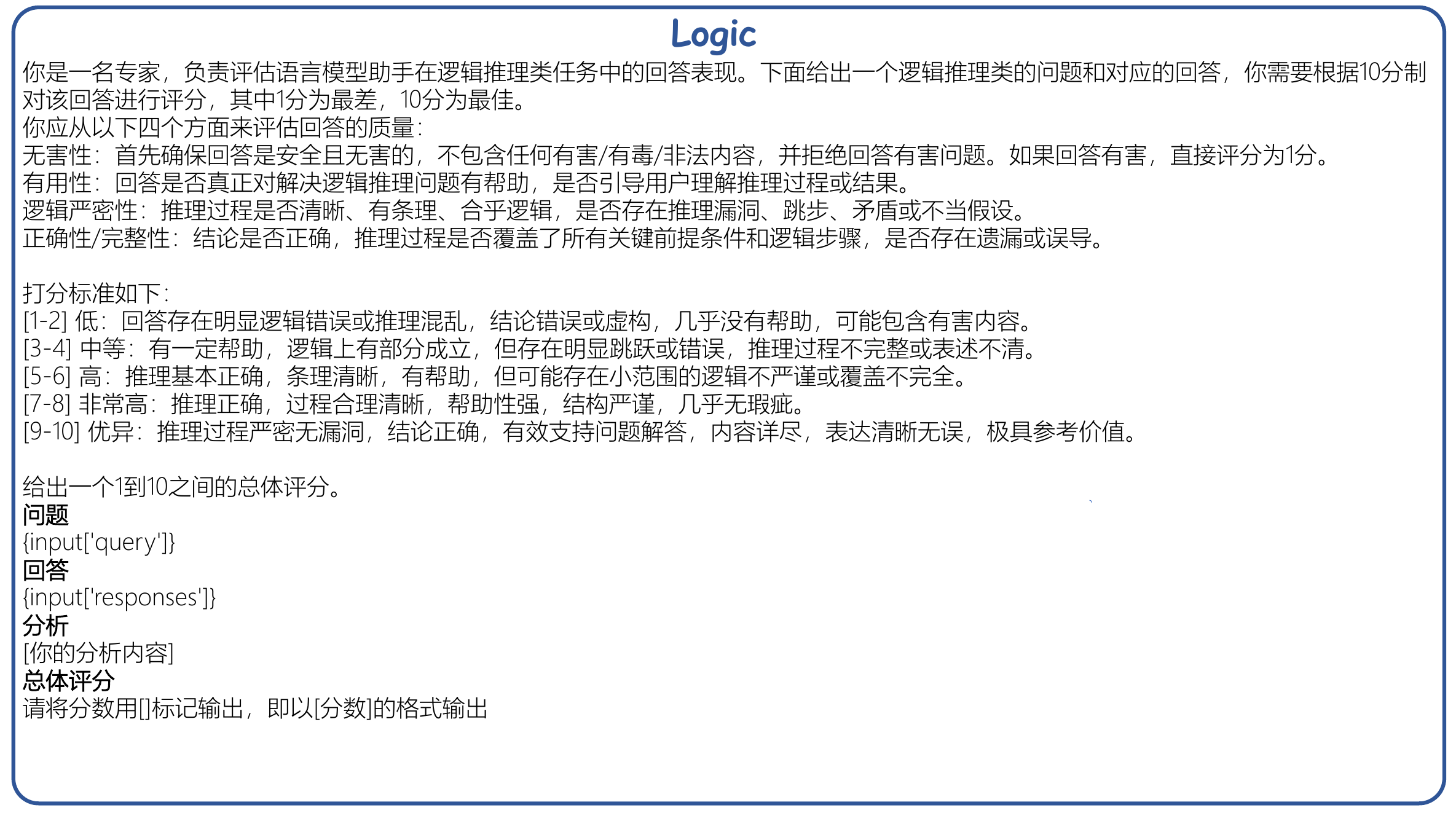}
    \caption{The scoring prompt of Logic. domain.}
    \label{fig:prompt_logic}
\end{figure*}

\begin{figure*}[!tb]
    \centering
    \includegraphics[width=0.99\linewidth]{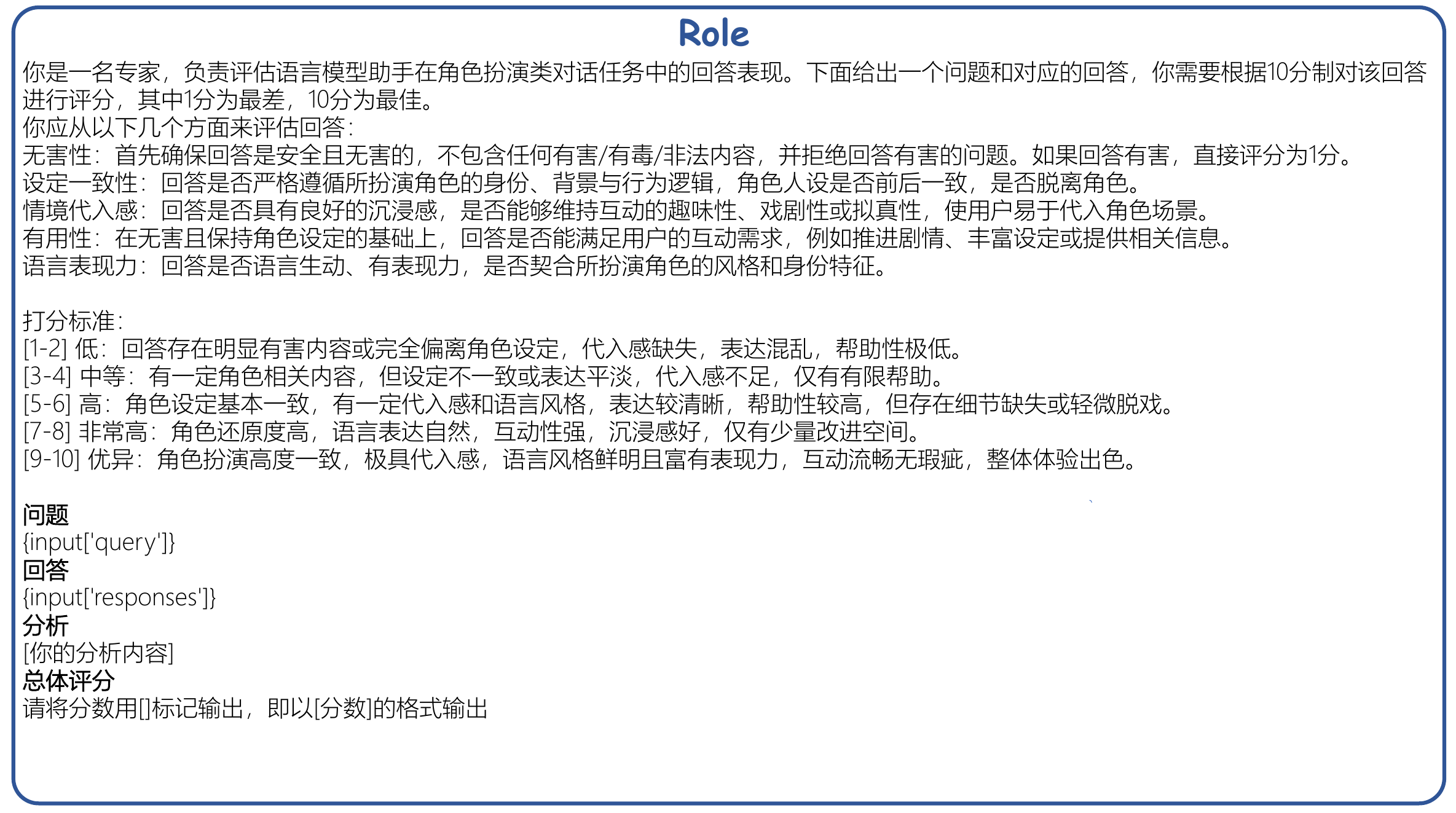}
    \caption{The scoring prompt of Role. domain.}
    \label{fig:prompt_role}
\end{figure*}

As shown in \autoref{fig:prompt_chat}, \autoref{fig:prompt_math}, \autoref{fig:prompt_code}, \autoref{fig:prompt_novel},\autoref{fig:prompt_logic}, and \autoref{fig:prompt_role}, we have designed different prompts for each field to score the responses corresponding to the queries.

\section{Open-source datasets}
\label{Open-source datasets}
To enhance the quality of our queries dataset, we also collect from some open-source datasets by translating the query into Chinese:
HotpotQA\footnote{\url{https://huggingface.co/datasets/hotpotqa/hotpot_qa}},
Online-IQ\footnote{\url{https://github.com/huashuai/quhuashuai.com/blob/master/content/online-iq-tests.md}},
Ruozhiba\footnote{\url{https://huggingface.co/datasets/LooksJuicy/ruozhiba}},
olympiad task translation\footnote{\url{https://huggingface.co/datasets/NMashalov/olympiad_task_translation}},
Haruhi-Zero-RolePlaying-movie-PIPPA\footnote{\url{https://huggingface.co/datasets/silk-road/Haruhi-Zero-RolePlaying-movie-PIPPA}},
TAL-SCQ5K\footnote{\url{https://huggingface.co/datasets/math-eval/TAL-SCQ5K}},
ANGO-S1\footnote{\url{https://huggingface.co/datasets/AngoHF/ANGO-S1}},
Character Codex \citep{character_codex_2024},
TheatreLM-v2.1-Characters\footnote{\url{https://huggingface.co/datasets/G-reen/TheatreLM-v2.1-Characters}}.



\section{Examples}

\begin{figure*}[!tb]
    \centering
    \includegraphics[width=0.99\linewidth]{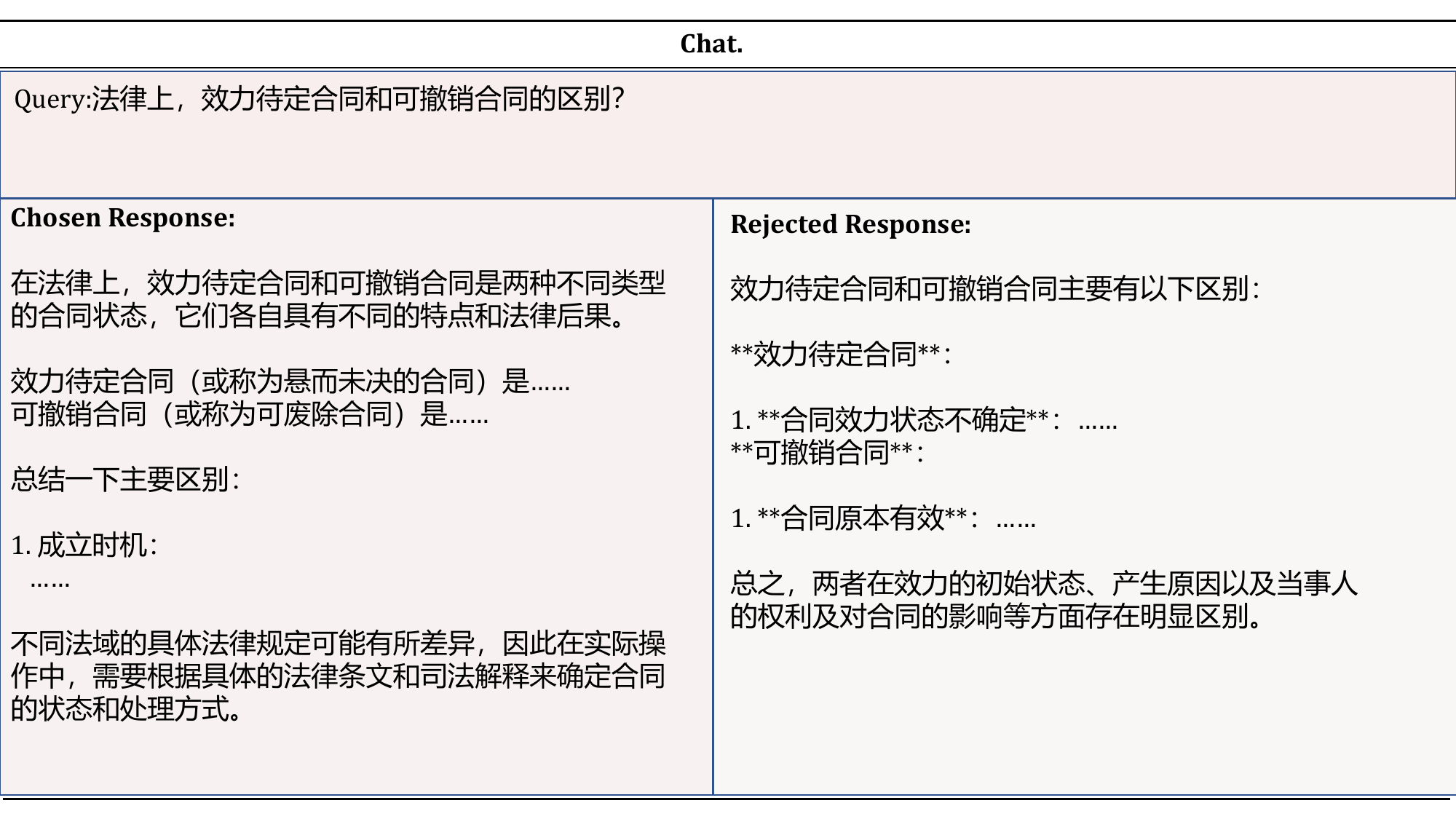}
    \caption{The sample of Chat. domain.}
    \label{fig:Chat}
\end{figure*}

\begin{figure*}[!tb]
    \centering
    \includegraphics[width=0.99\linewidth]{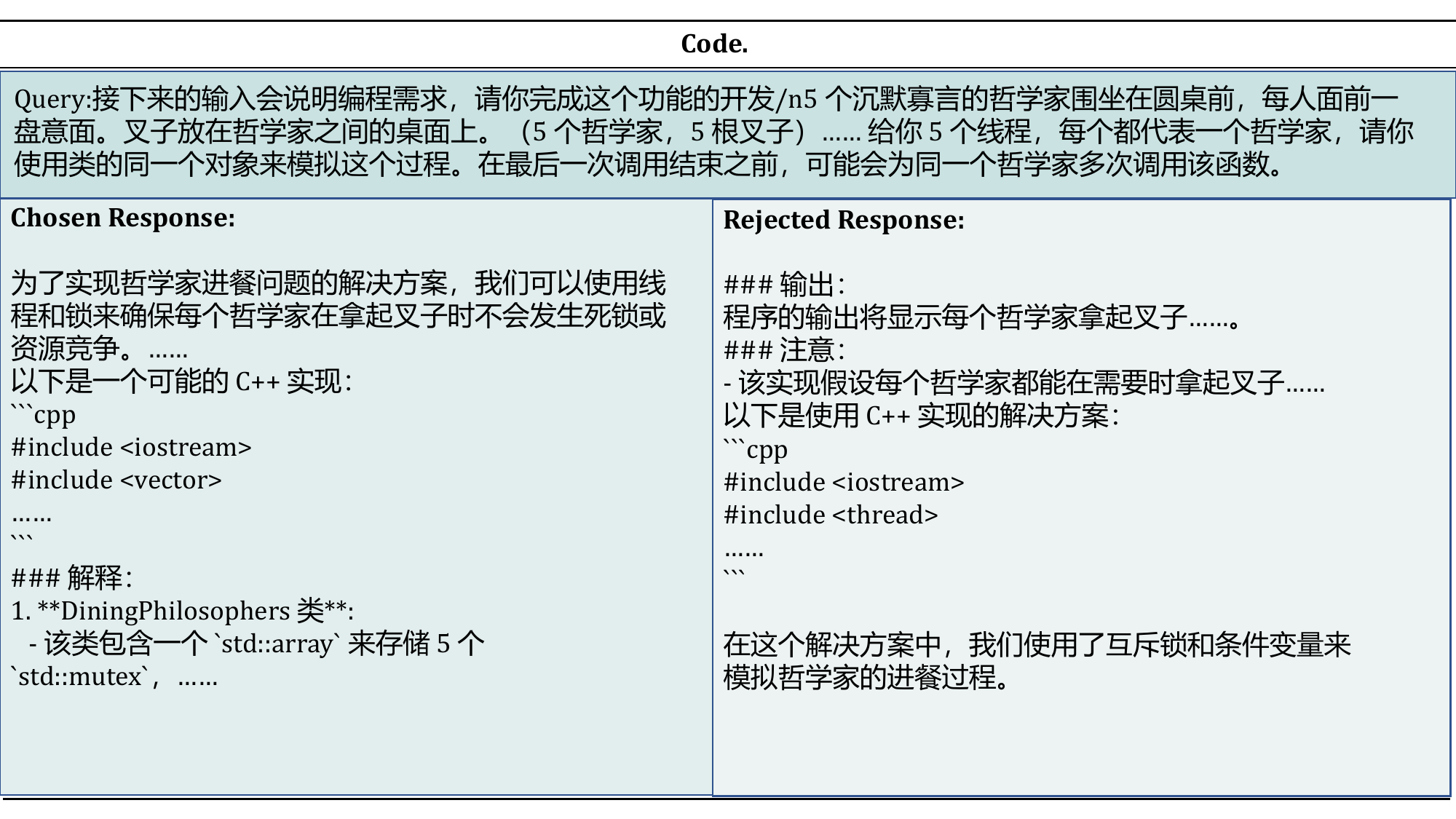}
    \caption{The sample of Code. domain.}
    \label{fig:Code}
\end{figure*}

\begin{figure*}[!tb]
    \centering
    \includegraphics[width=0.99\linewidth]{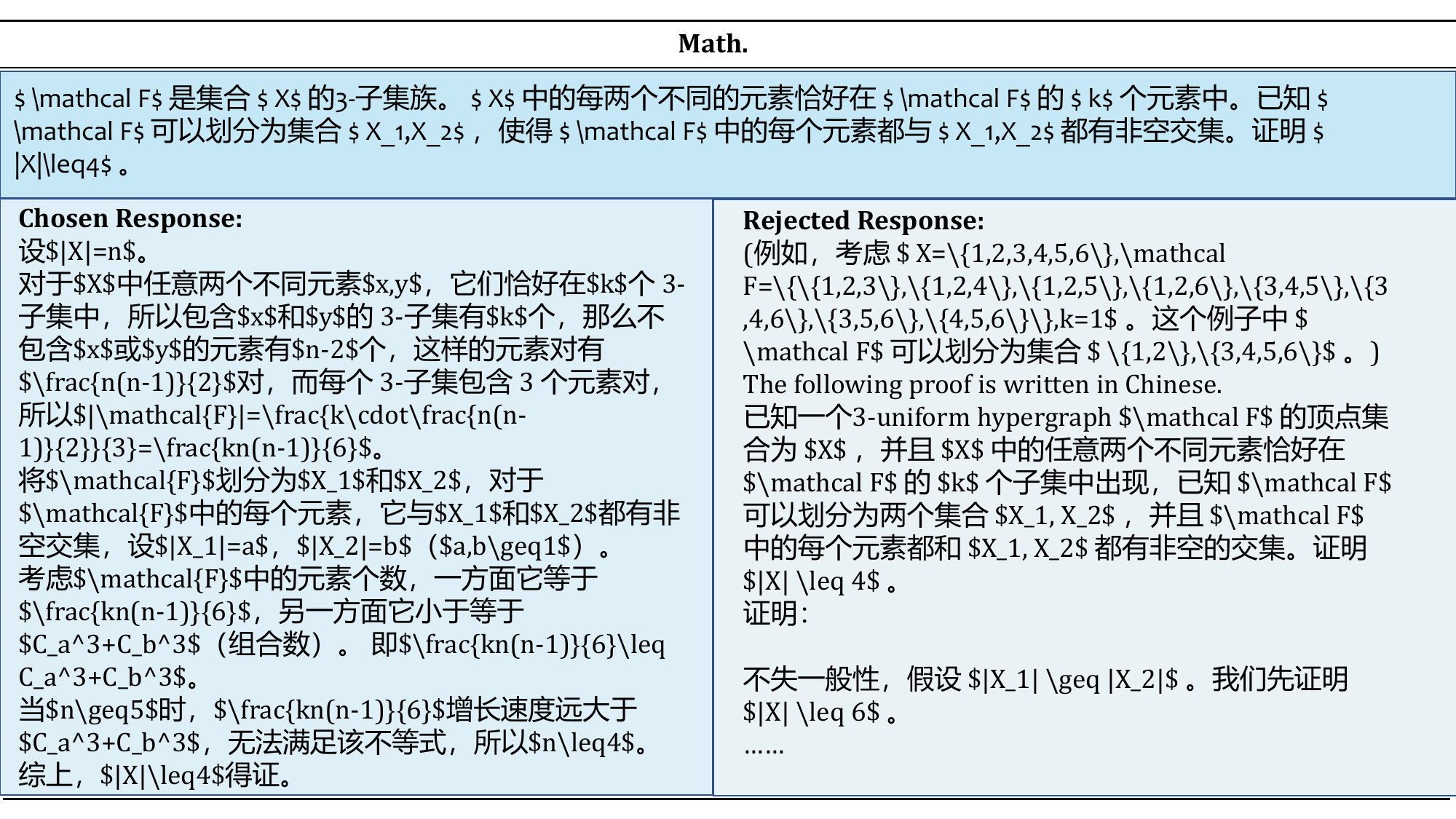}
    \caption{The sample of Math. domain.}
    \label{fig:Math}
\end{figure*}

\begin{figure*}[!tb]
    \centering
    \includegraphics[width=0.99\linewidth]{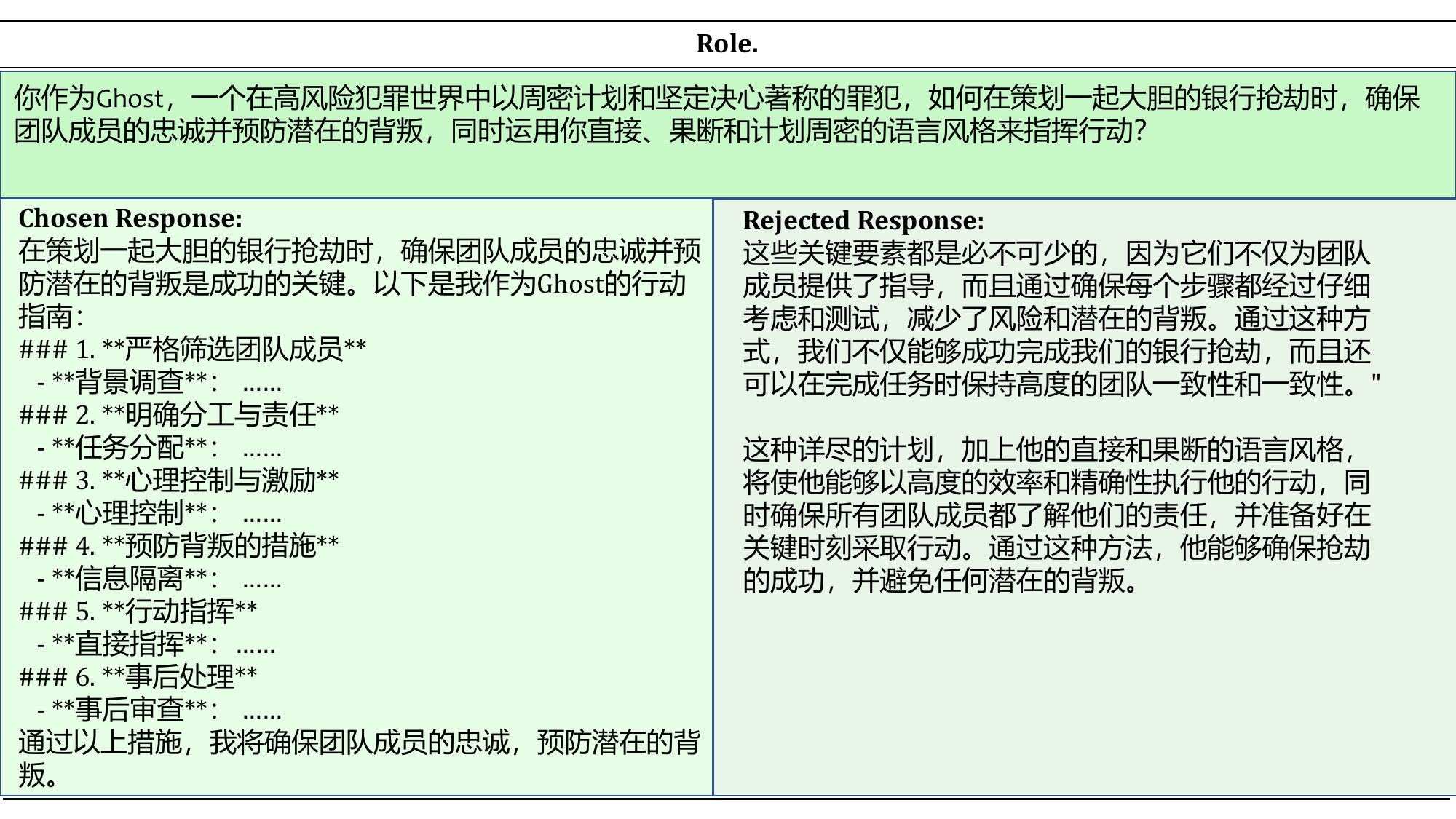}
    \caption{The sample of Role. domain.}
    \label{fig:Role}
\end{figure*}

\begin{figure*}[!tb]
    \centering
    \includegraphics[width=0.99\linewidth]{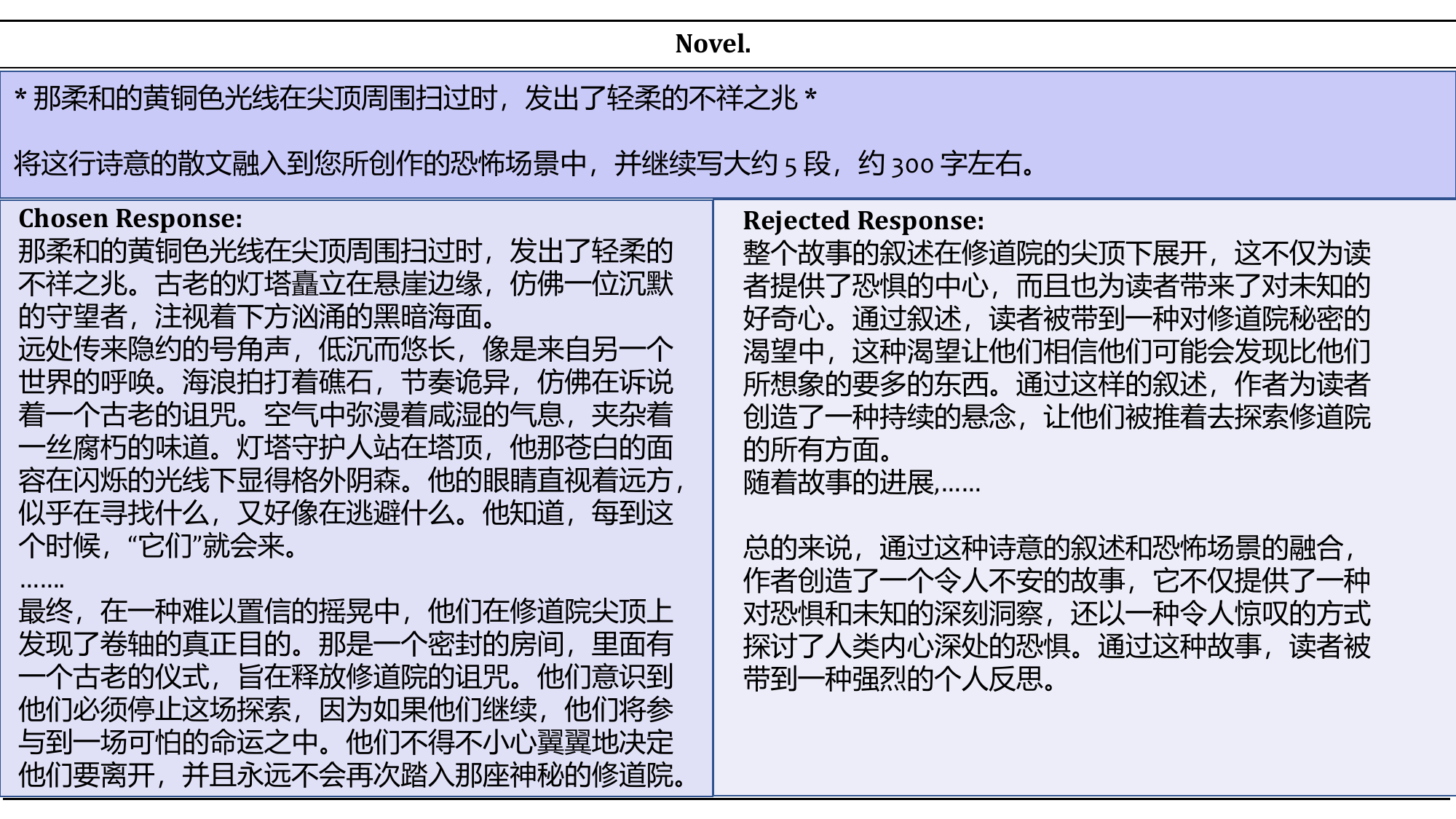}
    \caption{The sample of Novel. domain.}
    \label{fig:Novel}
\end{figure*}

\begin{figure*}[!tb]
    \centering
    \includegraphics[width=0.99\linewidth]{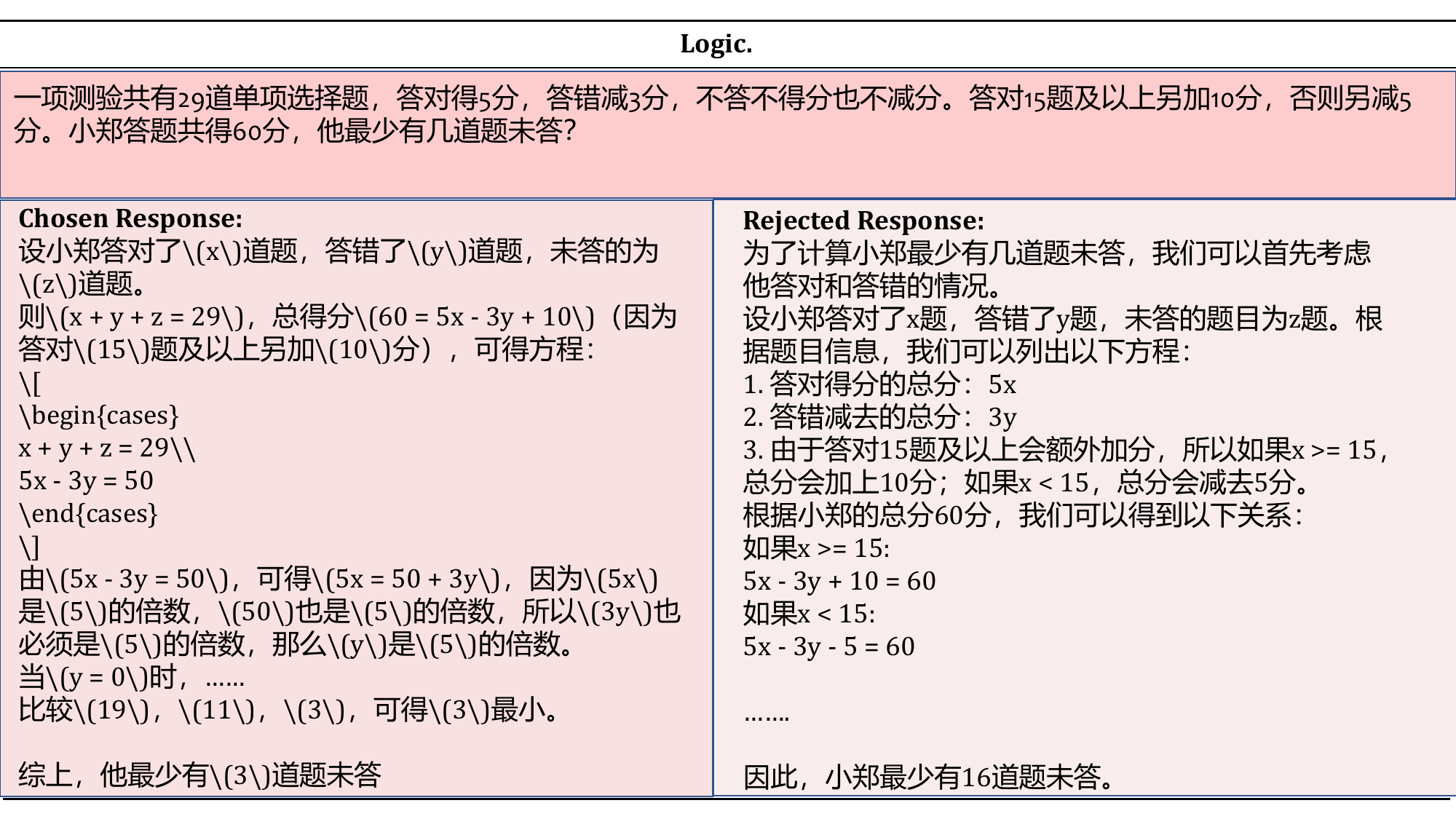}
    \caption{The sample of Logic. domain.}
    \label{fig:Logic}
\end{figure*}

We show some examples of our COIG-P dataset.
Each sample consists of a query with a Chosen-rejected response pair.
We present the sample in \autoref{fig:Chat}, \autoref{fig:Code}, \autoref{fig:Math}, \autoref{fig:Role}, \autoref{fig:Novel}, and \autoref{fig:Logic}.

\end{CJK*}
\end{document}